\documentclass[journal,article,submit,moreauthors,pdftex]{mdpi}

\usepackage{graphicx}
\usepackage{xcolor}
\usepackage{epsf} 
\usepackage{psfrag}
\usepackage{comment}
\usepackage{subcaption}
\usepackage{rotating}
\usepackage{array}
\usepackage{color,soul}

\firstpage{1} 
\pubvolume{xx}
\issuenum{1}
\articlenumber{5}
\pubyear{2019}
\copyrightyear{2019}
\history{Received: date; Accepted: date; Published: date}

\Title{Applications of machine learning \& IoT for Outdoor Air Pollution Monitoring and Prediction: \newline A Systematic Literature Review}

\Author{Ihsane Gryech  $^{1,2,}$*, Chaimae Asaad $^{1,3}$ , Mounir Ghogho$^{1,\ddagger,}$* and Abdellatif Kobbane $^{3}$}

\AuthorNames{Firstname Lastname, Firstname Lastname and Firstname Lastname}

\address{%
$^{1}$ \quad International University of Rabat, TICLab Research Laboratory, Morocco; \\
$^{2}$ \quad  ESAT-WaveCoRE, Department of Electrical Engineering, KU Leuven, Belgium;\\
$^{3}$ \quad ENSIAS, Mohammed V University in Rabat, Morocco; }

\secondnote{University of Leeds, School of IEEE , United Kingdom}
\corres{Correspondence: gryech.ihsane@gmail.com; (I.G.)}

\abstract{According to the World Health Organization (WHO), air pollution kills seven million people every year. Outdoor air pollution is a major environmental health problem affecting low, middle, and high-income countries. In the past few years, the research community has explored IoT-enabled machine learning applications for outdoor air pollution prediction. The general objective of this paper is to systematically review applications of machine learning and Internet of Things (IoT) for outdoor air pollution prediction and the combination of monitoring sensors and input features used. Two research questions were formulated for this review. 1086 publications were collected in the initial PRISMA stage. After the screening and eligibility phases, 37 papers were selected for inclusion. A cost-based analysis was conducted on the findings to highlight high-cost monitoring, low-cost IoT and hybrid enabled prediction. Three methods of prediction were identified: time series, feature-based and spatio-temporal. This review's findings identify major limitations in applications found in the literature, namely lack of coverage, lack of diversity of data and lack of inclusion of context-specific features. This review proposes directions for future research and underlines practical implications in healthcare, urban planning, global synergy and smart cities.}


\begin{document}



\section{Introduction}

Among the environmental threats that seriously harm human health and the ecosystem is air pollution\cite{first}. Major pollutants associated with public health include particulate matter (PM), carbon monoxide ($\text{CO}$), ozone ($\text{O}_{3}$), nitrogen dioxide ($\text{NO}_{2}$) and sulfur dioxide ($\text{SO}_{2}$)\cite{1WHO}. The World Health Organization (WHO) reports that 92 percent of the world's population lives in areas with polluted air, which causes 11.6 percent of global deaths\cite{1WHO}. In low- and middle-income countries, air quality is particularly poor due to urbanization and economic development, which cause growing levels of pollution\cite {WHO2}. Geographically, the highest annual average exposure levels of $\text{PM}_{2.5}$ were seen in Asia, Africa, and Middle East as listed in the State of the Air 2020 Report \cite{2020Report}. 

Air pollution can be approached from three perspectives: (i) measuring and monitoring air quality, (ii) analyzing air quality, and (iii) forecasting air quality\cite{Atmosphere}. Governmental environmental agencies are generally behind most deployments of air quality monitoring stations and therefore a verified source of air quality measurements \cite{Airparif1}. These stations are indeed precise and accurate, but are also very costly and require ongoing maintenance\cite{Tracing}. This has increased the interest in creating several low-cost and portable lightweight air quality sensor nodes, which are less costly, and cover a broader geographical area and contribute to the development of the field\cite{Tracing}. 

In order to better understand, assess, and control air quality and its impact on human health and the environment, air quality analysis and modeling are needed. The analysis and modeling are driven by an in-depth apprehension of sources of pollution, both natural and man-made, as well as interactions between emissions, atmospheric concentrations, meteorological features, and context-specific factors. This modeling is crucial for simulating the physical and chemical processes that affect air pollution dispersal and interaction in our atmosphere\cite{ZHANGG}.

Reducing harmful air pollutants entails both the identification of the sources causing pollution, and the development of continuous strategies for mitigating them.
Based on inputs of emission rates, urban climate, meteorological conditions, traffic related information, topography, and other factors affecting air quality, most models calculate and predict physical processes occurring within the atmosphere \cite{Josee}. 

The main approaches to air pollution prediction and forecasting are statistical linear methods and machine learning methods. However, a few works such as Hsieh et al.\cite{Statics} state that statistical linear methods do not consistently provide good estimations due to the complexity and variation in time-series data. In many scenarios, machine learning techniques have been shown to resolve issues of complexity \cite{MLL}.
The most widely used machine learning methods proposed for air pollution prediction and forecasting include Linear Regression, Support Vector Regression, Decision Trees, Random Forest, Gradient Boosting Regression, and ANN Multi-Layer Perceptron. 

With the growth of the machine learning field and the rising interest in air quality, more works have started tackling air pollution management by leveraging artificial intelligence and other technologies. A scientometrics and content analysis was conducted by Li et al. \cite{Related2} on the use of artificial intelligence techniques in the field of air quality forecasting. This study found that China holds the largest number of publications and citations, while the United States has a key role in the international cooperation network. Using such insights, Li et al. \cite{Related2} synthesized a number of suggested future directions and proposed the integration of forecasting results in decision-making processes for a better design of prevention and control plans.
Rybarczyk et al. \cite{Related1} performed a systematic review of the most relevant machine learning studies used in atmospheric pollution research. The review presented by Rybarczyk et al. \cite{Related1} used Scopus to search journal papers spanning the 2010–2018 time period and covering a range of topics related to machine learning and deep learning techniques used in atmospheric pollution estimation. The study excluded papers pertaining to physical sensors, and limited their search to journal papers. 
This review differs from existing work in a number of aspects. We present a technical perspective to the review of literature on monitoring and prediction of air pollution, with a specific focus on physical sensors. A method-oriented and cost-based categorization is used to present the selected literature.
Additionally, we include more recent contributions spanning the 2015-2022 period, with a scope encapsulating machine learning applications enabled by sensors of different types, namely low-cost sensors, high-cost monitoring stations and hybrid methods. Journal papers, conference papers and grey literature were reviewed in order to curb publication bias and report more transparent results. Other aspects to highlight the contributions of this paper include the research questions approached as well as the databases uses. We took into consideration publication bias, and conducted a thorough quality assessment process. The overall contributions, implications, and recommendations we present highlight new angles and perspectives that have not been presented in other reviews. 

This systematic literature review aims to analyze recent studies on the applications of machine learning techniques and IoT (Internet of Things) technology for air pollution prediction and forecasting. The scope of this paper includes different types of IoT-enabled monitoring applied in different countries with unique demographic, geographic, and economic conditions for outdoor air pollution prediction and forecasting.  

Various air pollutants are taken into consideration, while a cost-based and model-based classification is proposed. The methodology for this systematic review follows the PRISMA (Preferred Reporting Items for Systematic Reviews and Meta-Analyses) guidelines.

The remainder of the paper is structured as follows. Materials and methods including research questions, data sources and inclusion criteria are presented in \textit{Section }\ref{methods}. A summarization of the findings of the review, as well as a detailed comparison of the methods deployed in the included literature, are synthesized in \textit{Section }\ref{results}. Identified gaps, directions for future research as well as practical implications are presented in \textit{Section }\ref{discussion}, and conclusions are drawn in \textit{Section }\ref{conclusion}.



\section{Materials and methods}\label{methods}

This study is a Systematic Literature Review (SLR) conducted and reported according to the preferred reporting items outlined in the PRISMA (Preferred Reporting Items for Systematic Reviews and Meta-Analysis) statement\cite{prisma}.

\begin{figure}[!htb]
    \centering
    \includegraphics[width=0.8\textwidth]{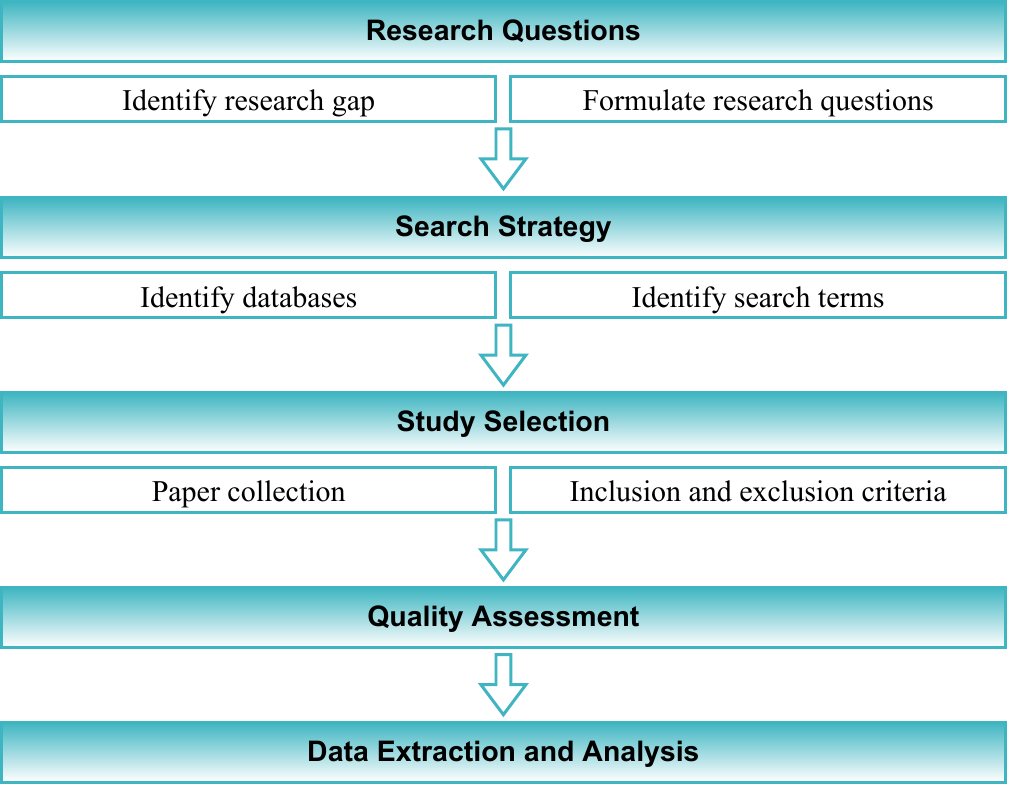}
    \caption{PRISMA Methodology followed in the systematic review process}
    \label{fig:methodology}
\end{figure}

\begin{figure}[H]
    \centering
    \includegraphics[width=0.8\textwidth]{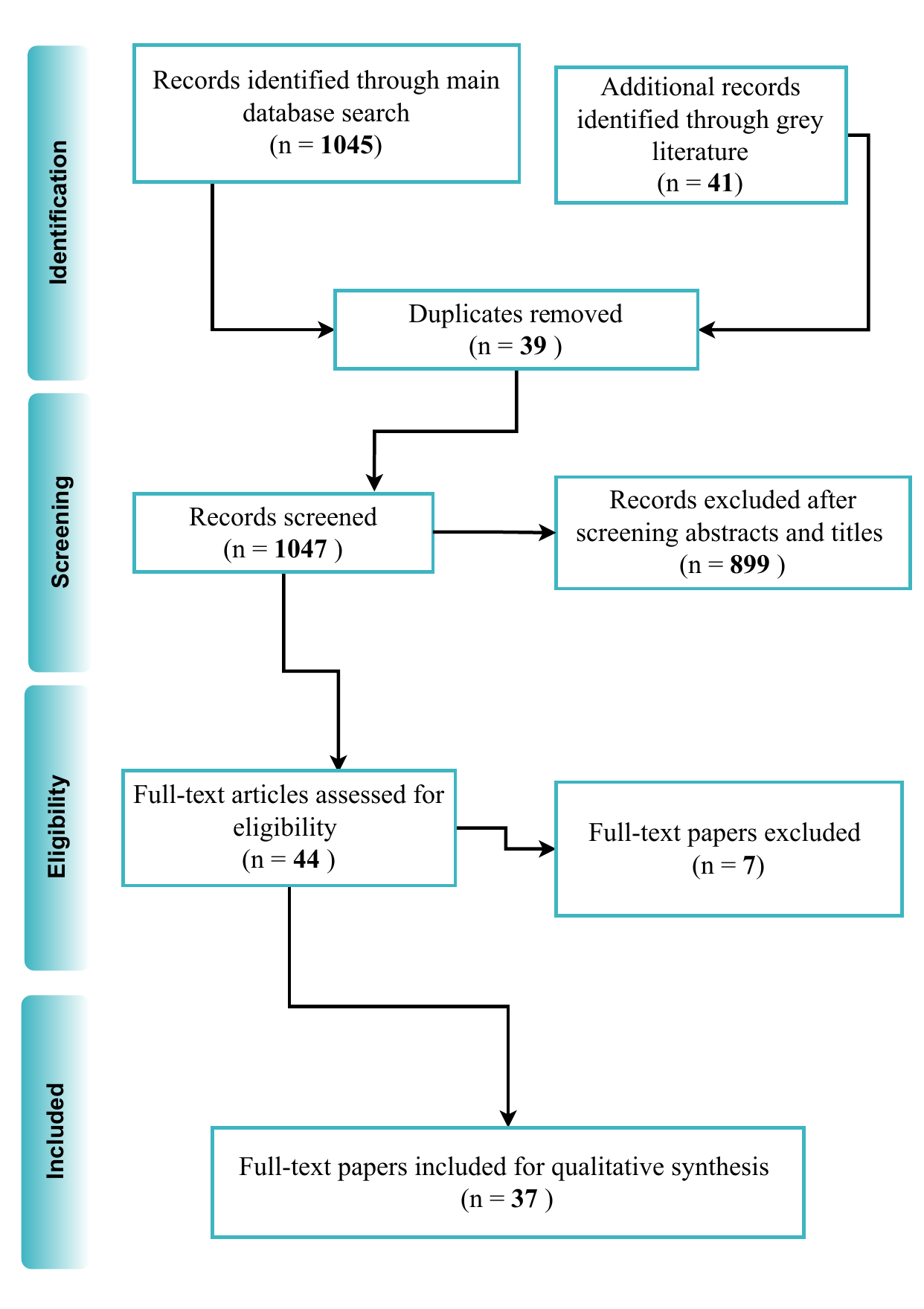}
    \caption{Flow Diagram for the selection of the literature reviewed. \textit{The abstract screening process resulted in 44 studies identified for detailed review of full-text articles. After this review, we further excluded studies that did not meet the inclusion criteria, or that did not answer the research questions. We identified a total of 37 studies that met our eligibility criteria and addressed the research questions.}}
    \label{fig:flowchart}
\end{figure}

This paper followed the SLR process (Fig. \ref{fig:methodology}) comprised of three main phases: (i) Identification, (ii) Screening and (iii) Eligibility. 
Upon the formulation of research questions(s), the Identification phase defined the search strategy, including the choice of data sources and extraction methods to be used for the collection of relevant papers. Inclusion and exclusion criteria aligned with the specific requirements and scope of the review were outlined in the screening and eligibility phases, where papers were filtered based on titles and abstracts (screening) and full-text (eligibility). Fig. \ref{fig:flowchart} details the results of each phase of the PRISMA process. The answers to the research question(s) were then synthesized, while the challenges, opportunities, and limitations were summarized. The results of each of these steps are detailed in the subsequent sections below.

\subsection{Research Questions}

According to the United States Environmental Protection Agency (EPA)\footnote{\url{https://www.epa.gov/report-environment/outdoor-air-quality}}, outdoor air pollution is associated with a number of human health effects including heart attacks, asthma attacks, bronchitis, hospital and emergency room visits, restricted activity days and premature mortality. Therefore, it is paramount to leverage and apply various technologies such as machine learning and Internet of Things for air pollution monitoring and forecasting. 

In this his systematic review, we identified essential research questions and tried to find relevant answers through a detailed study. Hence, the research questions devised for this systematic review are:

\begin{itemize}
    \item \textbf{RQ1:} What are the machine learning techniques used for outdoor air pollution prediction and forecasting?
    \item \textbf{RQ2:} What are the combinations of monitoring sensors and input features used for outdoor air quality prediction and forecasting? 
\end{itemize}

These questions have been conceptualized to achieve the main contribution of this paper: a systematic review of machine learning and monitoring methods used for air pollution forecasting and prediction. Moreover, this paper aims to provide a comprehensive synthesis of the methods, pollutants, and models used, as well as the deployment strategies and numbers of sensors utilized. This paper also gives an insight into the state of research in the field, in addition to the main challenges and potential practical implications.  

\subsection{Search Strategy}
To address the research questions, the authors have used different databases: IEEE Xplore, Google Scholar, ACM Digital Library, SpringerLink, Science Direct, MDPI and the grey literature resource Arxiv. The search for relevant publications was initiated on June 1st, 2021.

Using the research questions, keywords, and queries were formulated according to the requirements of each scientific database used in the search. Queries used a combination of the keywords "\textit{Machine Learning}", "\textit{Outdoor Air Pollution}", "\textit{Sensor}", "\textit{Monitoring}". All searches used the Boolean operator "AND" as well as the parameters "full-text" and "all metadata" when such specifications were supported by the search database. After completing the first draft of search strings, we examined the results of each search string against each database to check the effectiveness of the search strings. The search time frame covered publications spanning the January 2015 - January 2022 period.

The results of the initial search are reported in Table \ref{tab:numberofpapers}. After the paper screening, exclusion, and duplicates removal, 148 papers remained. These studies were further processed according to the inclusion and exclusion criteria.

\renewcommand{\arraystretch}{1.25}
\begin{table}
\caption{Number of Papers Collected from each Database}
\centering
\begin{tabular}{|c|c|}
\hline
\textbf{Database} & \textbf{Number of Papers Collected} \\
\hline
IEEE Xplore & 345 \\
Google Scholar & 124 \\
MDPI & 45 \\
ACM Digital Library & 477 \\
SpringerLink & 16\\
ScienceDirect & 38\\
Arxiv & 41  \\
\hline
\end{tabular}
\label{tab:numberofpapers}
\end{table}

\subsection{Study Selection and Data Extraction}
The collected papers from the initial search were screened according to the pre-set inclusion and exclusion criteria (Table \ref{tab:inclusionexclusion}). The paper selection process consisted of 2 phases. First, based on the inclusion and exclusion criteria, the papers were independently screened by two researchers through title and abstract screening. The papers selected in this phase were then assessed through full-text screening. The authors cross-checked the selection results and resolved any disagreement on the selection decisions. All disagreements in either the first or the second phase were resolved by consensus, and a third researcher was consulted to finalize the decision.
The researchers involved in this selection are familiar with the systematic process of paper selection, and have conducted systematic reviews before. In terms of expertise, the authors involved in the paper selection have expertise in machine learning, air pollution monitoring and the application of machine learning for air pollution prediction. The third researcher involved in conflict resolution has senior expertise in these domains.
The inclusion and exclusion criteria were formulated by the authors to effectively select relevant papers. The documents were analyzed to address the different methods related to the intersection of applications using sensors and machine learning in order to predict and forecast air quality.

Both journal papers and conference papers written in English and within the scope of the research questions were included. Commercial papers, letters to the editor, e-books, books and PhD dissertations were excluded. Papers were excluded if they were review papers or only presented a theoretical framework without a tangible application. Papers presenting AQI (Air Quality Index) prediction were excluded since AQI is based on measurement of PMs, Ozone, Nitrogen, Sulfur Dioxide and Carbon Monoxide emissions. Therefore, AQI gives a score corresponding to a level of air pollution on a country-level scale \cite{aqidiff}. The authors deemed that the aim of the review was to identify applications focusing on the prediction of particular pollutants rather than the prediction of the AQI.

The relevant data was extracted from the selected publications for further analysis. In order to conduct this systematic literature review, the following information was extracted:

\begin{itemize}
    \item Author list,
    \item Titles and abstracts,
    \item Year of Publication, 
    \item Associated Database, 
    \item Source of Data, pollutants, method of prediction, machine learning model(s) used,
    \item Number of sensors used and sensor deployment strategy.
\end{itemize}
\renewcommand{\arraystretch}{1.25}
\begin{table}[]
\caption{Inclusion and Exclusion Criteria for Paper Selection}
\begin{tabular}{|
>{\columncolor[HTML]{FFFFFF}}p{7cm} |
>{\columncolor[HTML]{FFFFFF}}p{7cm} |}
\hline
\textbf{Inclusion Criteria}                                         & \textbf{Exclusion Criteria}               \\ \hline
Within scope of research questions &
Does not relate to prediction of outdoor air pollution \newline Does not include use of sensors or monitoring stations \newline Focuses on Air Quality Index \\ \hline
Within time frame of publication: January 2015-January 2022 & -                                \\ \hline
Includes tangible application  & Presents a theoretical framework \\ \hline
Is a journal or conference paper &
Book, e-book, letter to editor, magazine, abstracts, case reports, comments, reviews, posters \\ 
\hline
Written in English & Other languages \\ 
\hline
\end{tabular}
\label{tab:inclusionexclusion}
\end{table}

\subsection{Risk of Bias and Quality Assessment}

This systematic literature review followed the PRISMA guidelines to screen and select the relevant literature. A few limitations contributing to risk of bias are worth highlighting. The selection of keywords to use in the initial query search can be influenced by bias. Additionally, the eligibility criteria defined by the authors are prone to subjectivity and thus increase risk of bias. However, based on the PRISMA guidelines, the best possible criteria were followed to complete this systematic literature review. Grey literature sources were included in the initial data sources to curb publication bias. Independent selection processes and disagreement resolution techniques were utilized to ensure that the selection of publications is both transparent and relatively objective. A quality assessment system based on consensus between authors was used to ensure that publications of high quality and substantial contributions were included. This system was based on a checklist of the following criteria:

\begin{itemize}
    \item Are the methods used clearly defined and applied ? 
    \item Are the methods applied successfully and correctly? 
    \item Are accuracy values and efficiency/confidence levels reported?
    \item Do the contributions outweigh the limitations of the study?
\end{itemize}

\subsection{Characteristics of selected papers}
\begin{figure}[H]
\centering
\begin{subfigure}{.5\textwidth}
  \centering
    \includegraphics[width=1\textwidth]{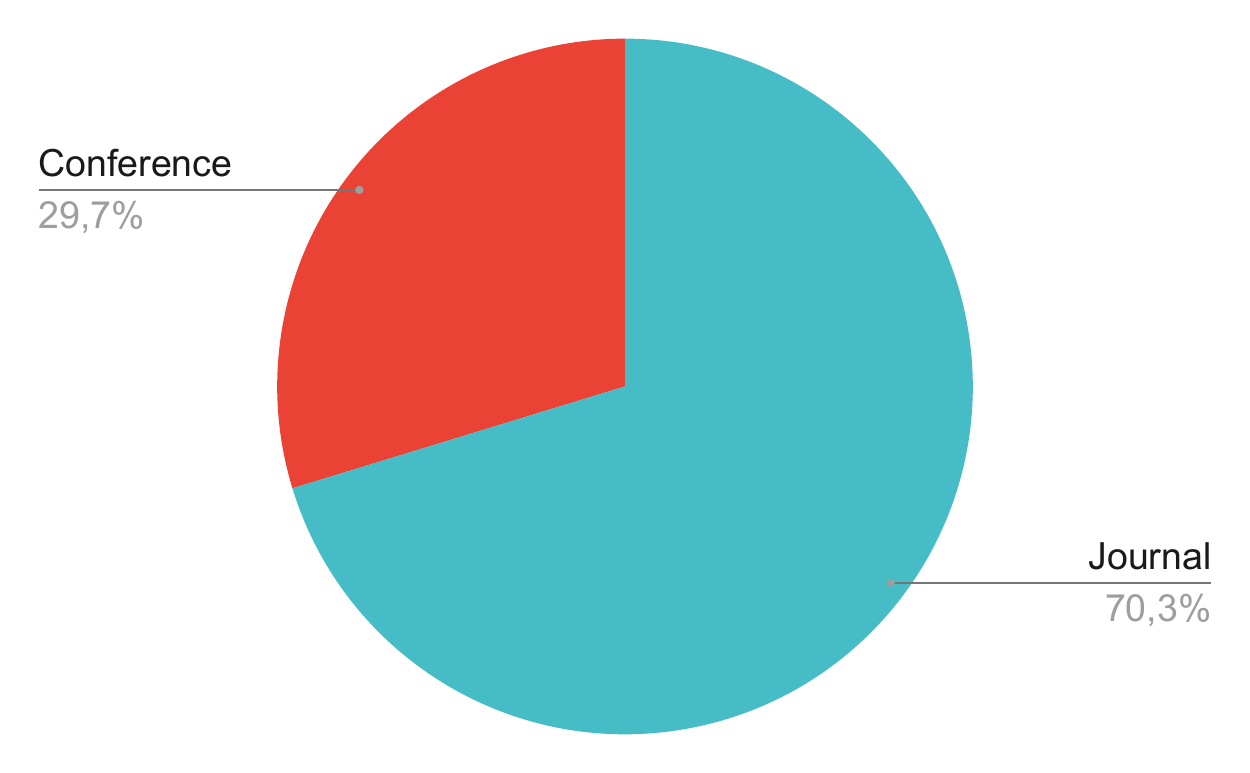}
    \caption{}
    \label{fig:ByType}
\end{subfigure}%
\begin{subfigure}{.5\textwidth}
  \centering
   \includegraphics[width=1\textwidth]{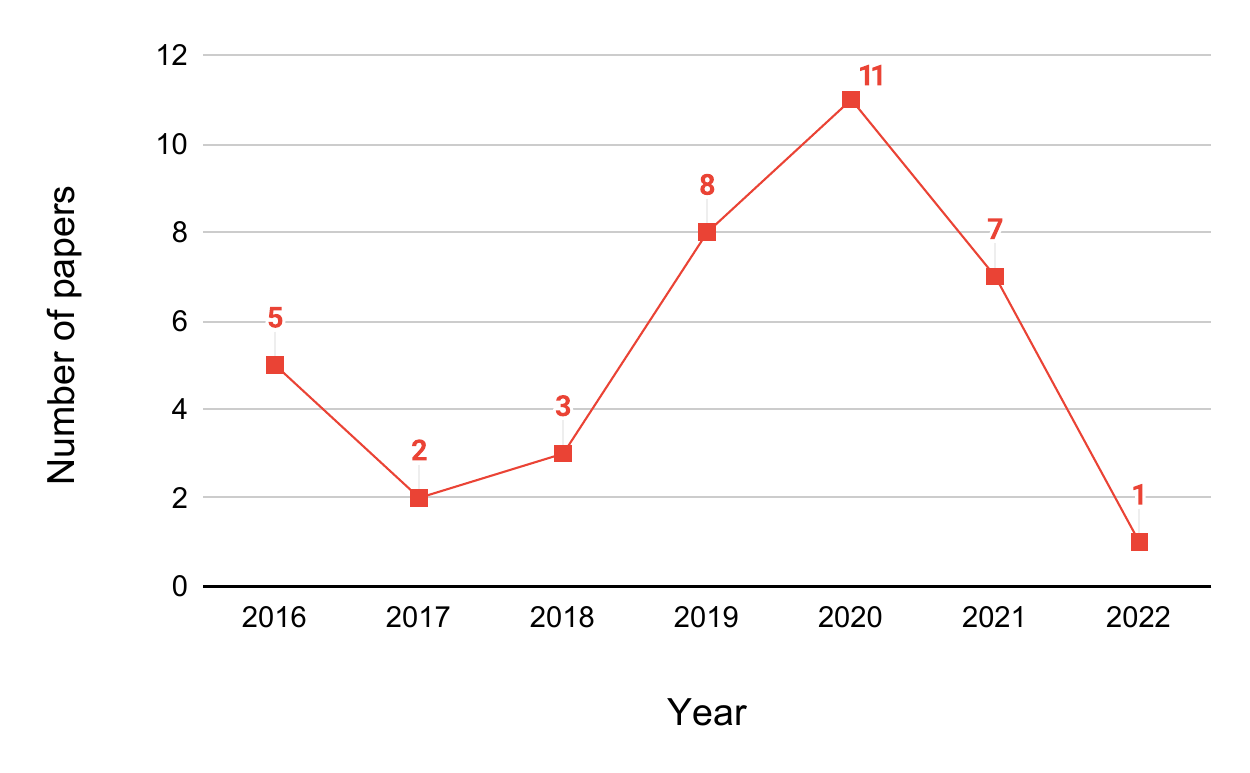}
    \caption{} 
    \label{fig:ByYear}
\end{subfigure}
\caption{Distribution of included papers by (a) publication type and by (b) publication year}
\label{fig:ByType}
\end{figure}

The search process resulted in a total of \textit{1086} articles distributed over both the main and grey databases used. After the removal of duplicates, \textit{1047} papers remained. Of these, \textit{899} studies were excluded after the title and abstract screening, as they did not fulfill the inclusion criteria. Of the \textit{44} studies that were full-text screened, \textit{7} did not meet the inclusion criteria. A total of \textit{37} studies were selected for inclusion in the current review, as summarized in the following sections. The selection process took 6 months to complete.

The papers included in the systematic literature review were distributed as follows: 70.3\% are journal papers and 29.7\% are publications of conference proceedings (Fig. \ref{fig:ByType}-(a)). After the two screening steps or throughout the duplicate elimination process, no works from grey literature remained.
The publications included span the 2015-2022 period. As can be seen in Fig. \ref{fig:ByType}-(b), the highest number of publications included were published in 2020. 


\section{Results}\label{results}

To address the formulated research questions, we conducted a cost-oriented categorization of monitoring and machine learning applications for predicting outdoor air pollution.

Machine learning prediction of air pollution can be enabled by low-cost, high-cost and hybrid sensing units. 
Low-cost air quality monitoring sensors are an equipment that can measure a variety of airborne parameters and contaminants while remaining affordable and available to a wide range of users. These sensors are made to be cost-effective, compact, and easy to deploy, allowing communities, businesses, and people with limited budgets to use them. Low-cost sensors are frequently small and lightweight, allowing for easy deployment in a variety of settings, such as homes, workplaces, and urban areas. Nevertheless, the accuracy and precision of these sensors are questionable, when using them for essential applications or regulatory compliance, limitations should be taken into account because they might not always match the accuracy of professional-grade monitoring equipment. Low-cost IoT sensors are often developed in research laboratories \cite{C15} using affordable materials. These are the IoT sensors we are considering in this work.


High-cost monitoring includes expensive monitoring stations often deployed by governmental environmental entities in order to regulate and monitor air quality and ensure residents safety. Although they are highly reliable and can measure a wider range of pollutants, maintenance is continuously needed, further adding to their financial burden. Due to this high cost, these monitoring stations are often scarce and deployed in specific areas and are not always available to developing and low-income countries \cite{r83}.

Hybrid monitoring combines accurate, reliable data from high-cost monitoring stations with data from low-cost sensor nodes. That guarantees both precise measurements of the air quality and a broader geographic coverage. This type of monitoring method is not as widely used since it requires access to data from both high and low cost sensors.

\subsection{Applications of low-cost IoT and ML for outdoor air pollution}

\begin{figure}[h]
    \centering
    \includegraphics[width=0.9\textwidth]{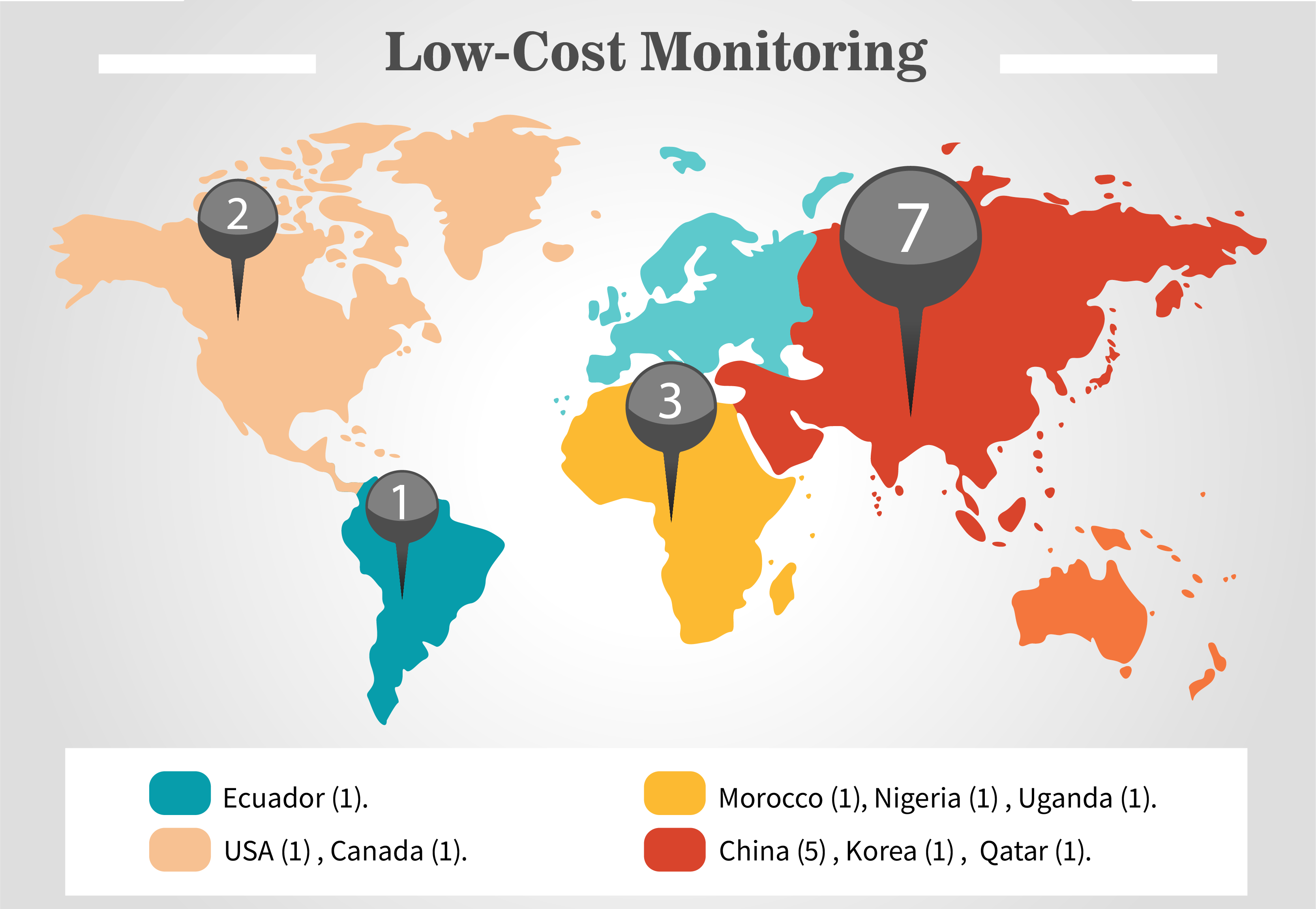}
    \caption{Geographical distribution of included papers using low-cost sensor nodes}
    \label{fig:Low-Cost Map}
\end{figure}

\begin{figure*}[h]
\centering
\includegraphics[width=\textwidth]{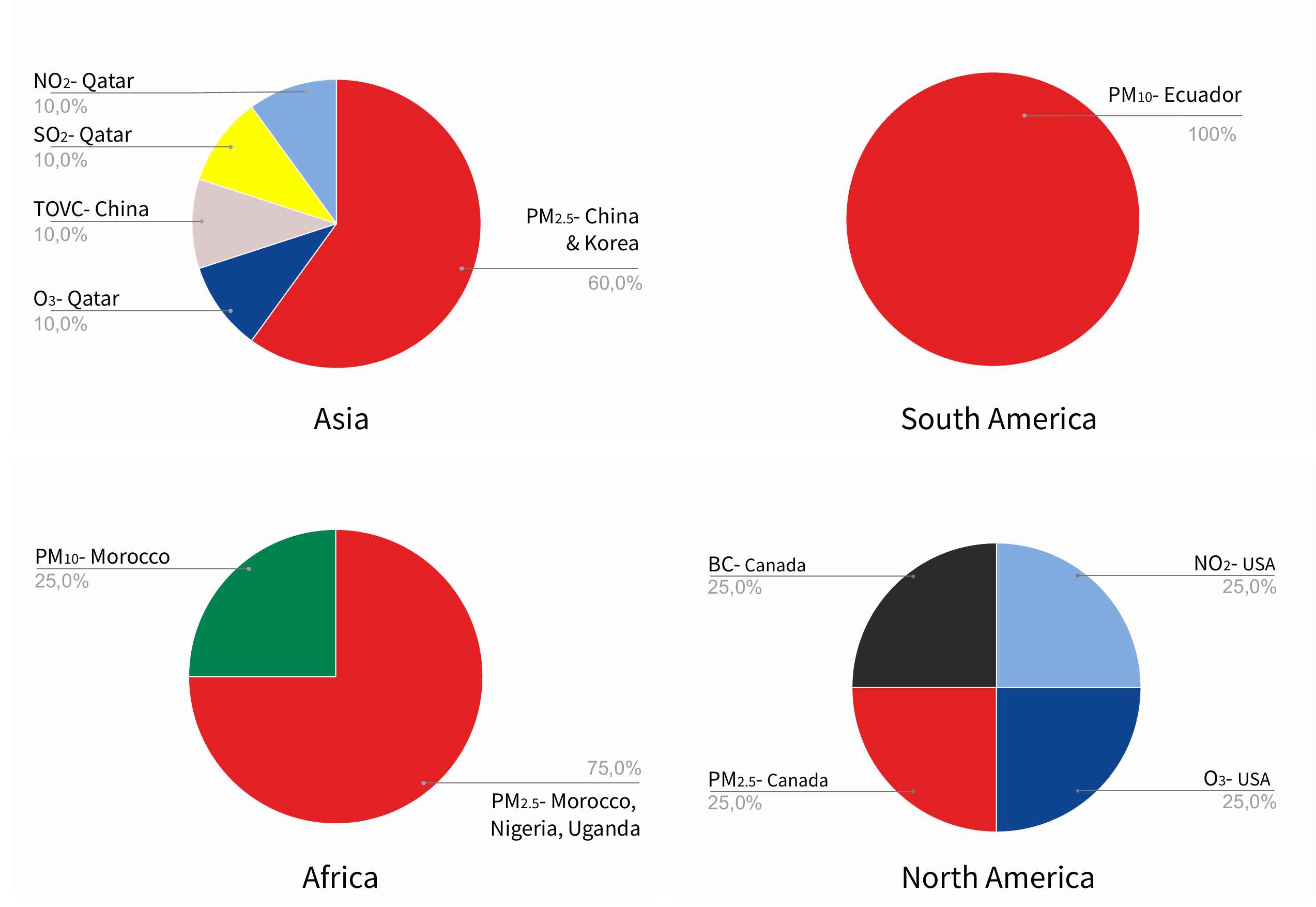}
\caption{Proportion of pollutants predicted in included literature using Low-cost enabled monitoring and machine learning for outdoor air pollution prediction. \textit{$\text{PM}_{2.5}$- Particulate matter with a diameter <2.5 microns, $\text{PM}_{10}$- Particulate matter with a diameter <10 microns, $\text{BC}$- Black Carbon, $\text{O}_{3}$- Ozone, $\text{NO}_{2}$- Nitrogen Dioxide,
$\text{SO}_{2}$- Sulfur Dioxide, and $\text{TOVC}$- Total volatile organic compounds.}}
\label{figLC}
\end{figure*}

Geographically, selected papers presenting applications using low-cost enabled monitoring and ML for outdoor air pollution prediction use data collected mostly from Asia, Africa, North America and South America, sequentially. Data from Europe was notably absent from selected papers, as can be seen in Fig. \ref{fig:Low-Cost Map}. A wider range of pollutants was predicted based on data from Asia and North America, while Africa and South America mostly focused on Particulate Matter (Fig. \ref{figLC}).
A summary of pollutants, geographical data sources, estimation methods, machine learning models and deployment strategies used in the included literature is illustrated in \textbf{Table \ref{fig:LowCostTable}}.

\subsubsection{Time series prediction}

Of the selected literature, two papers \cite{r68,r55} used time series for the forecasting of $\text{PM}_{2.5}$. LoRa\footnote{(\url{https://lora-alliance.org/})} (long-range wireless data communication technology), coupled with a cloud-based system model and a long short-term memory (LSTM) cyclic neural network was used in \cite{r68} to predict the next few hours’ $\text{PM}_{2.5}$ values and carry out an air quality index analysis of $\text{PM}_{2.5}$. Results showed that the time step of 48h resulted in the best performance with the lower RMSE (Root Mean Square Error) of 18.65. 
A similar LoRa-based wireless hardware was used in \cite{r55} for a short-term forecasting method based on autoregressive integrated moving average and vector autoregressive moving average (VARMA). The results showed that the overall root-mean-square error and correlation coefficient of the VARMA models integrated with hierarchical clustering are improved by 7.77\% and 3.7\%, respectively, compared with the usual single node-based forecast model.

\subsubsection{Feature-based prediction}
Several papers \cite{r30,r103,r42,r45,r75} conducted feature-based predictions to estimate $\text{PM}_{2.5}$ concentrations, while two papers \cite{r83,r112} focused on estimating $\text{PM}_{2.5}$, $\text{PM}_{10}$, and Black Carbone, respectively. $\text{O}_{3}$ and $\text{NO}_{2}$ concentrations were predicted in \cite{r7}.\\

Different deployment strategies were used in the selected literature. 151 mobile sensors fixed on taxis with random routes were deployed in Beijing, China \cite{r30}. Only one sensor node was deployed at one location in Lagos, Nigeria \cite{r103}. Air quality sensors built-in 500 vehicles were used to collect data in fine-grained spatial and temporal resolution all over the city of Beijing, China \cite{r42}. 
For cost-efficient data collection, 5 taxis with built-in low-cost sensor nodes and 1 customized hexacopter were implemented in Hangzhou, China \cite{r45}. One low-cost sensor node was used in Quito, Ecuador \cite{r83}. Three nomadic and 4 mobile low-cost sensor nodes were deployed in neighborhoods of different income levels in Rabat, Morocco \cite{r83}. Two low-cost sensor nodes were deployed on the rooftop of moving vehicles in Toronto, Canada  \cite{r112}. And a network of 31 low-cost sensors were used in South California \cite{r7}.

A Real-time Ensemble Estimation Model based on Gaussian Process Regression for air pollution of the unmonitored areas was presented in \cite{r30}, pivoting on the diffusion effect and the accumulation effect of air pollution. A two-layer ensemble learning framework and a self-adaptivity mechanism using nine features including $\text{PM}_{2.5}$ concentration, monitoring time, longitude, latitude, wind direction X, wind direction Y, temperature, pressure, and humidity. Results show an error varying between 0.1 and 0.2. ARIMA, Prophet, Exponential Smoothing, Neural Network Auto Regressive (NNAR), Support Vector Machine, XGBoost, and Random Forest were all trained to generate 24-hour-ahead forecasts of $\text{PM}_{2.5}$ concentration based on air quality data in \cite{r103}. Results showed ensemble models (XGBoost-RF-ARIMA) generated more reliable forecasts compared to standalone algorithms. However, the limited number of sensor nodes used in this study presents a major limitation, hindering the representativeness of the results. 

A crowdsourcing based urban air quality sensing system, along with Random Forest, were used in \cite{r42} to estimate the concentration value of $\text{PM}_{2.5}$ and TVOC (Total Volatile Organic compounds) of local environment. Results showed a 92.4\% accuracy in window status recognition and less than 9\% bias for estimating the air quality.
A three-dimensional (3D) spatial-temporal fine particulate matter monitoring system (BlueAer) complemented by a 3D probabilistic concentration estimated method (3D-PCEM) were used in \cite{r45} and compared to other estimation algorithms (Linear Interpolation, MLR, ANNA, GP). Results showed that the average estimated error of BlueAer was enhanced by 15.4\% and 41.0\% compared to Gaussian Process (GP) and Artificial Network (ANN) respectively. 
$\text{PM}_{2.5}$ concentrations from traffic were predicted in \cite{r75} by extracting data from Web-based applications such as Google Traffic and training a decision tree algorithm for different days using three variations of the model based on an Inverse Distance Weighted model (prediction accuracy 50\%–71\%), on traffic (prediction accuracy 61\%–71\%), and both traffic and time (6.5\% increase in prediction accuracy). 

A low-cost monitoring system (MoreAir) was proposed in \cite{r83} for feature-based prediction for both $\text{PM}_{2.5}$ and $\text{PM}_{10}$ using Random Forest, Support Vector Machine and Multiple linear Regression, and including both meteorological and traffic features as predictors. Random Forest outperformed Support Vector Machine and Multiple Linear Regression with an $R^2$ of 0.63 for $\text{PM}_{2.5}$ and 0.57 for $\text{PM}_{10}$.

$\text{PM}_{2.5}$ and Black Carbone concentrations were measured and estimated in \cite{r112} using sensors, a moving camera to collect real-time traffic and models developed using Land-use Regression (LUR) and other machine learning methods.
Only one selected work \cite{r7} predicted $\text{O}_{3}$ and $\text{NO}_{2}$ concentrations using land use data combined with hourly pollution measurements at high temporal resolution. By including all land use, altitude and traffic related predictors, the random forest model performed well for both pollutants concentrations with an $R^2$ of 0.93 for $\text{NO}_{2}$ and 0.73 for $\text{O}_{3}$.

\subsubsection{Spatio-temporal prediction}

Only a few works \cite{r114,r115,r130} included spatio-temporal forecasting. In Beijing, China, 260 electrical vehicles were equipped with low-cost sensors to collect real-time spatial data and infer the distribution of $\text{PM}_{2.5}$ concentrations using machine learning models \cite{r114}. 
Results showed that the gradient boosting model using the extreme gradient boosting (XGBoost) algorithm inferred the best spatial distribution of $\text{PM}_{2.5}$ emissions at 1x1 km resolution on an hourly average with a $R^2$ value of 0.8. 

More than 80 low-cost sensor nodes were deployed in \cite{r130} in both fixed locations and using mobile monitors on motorcycles and taxis in Uganda. 
The monthly averaged $\text{PM}_{2.5}$ measurements from 23 of these nodes were fed to eight ML algorithms, as well as ensemble modeling to predict monthly $\text{PM}_{2.5}$ concentrations. Results showed that the overall average $\text{PM}_{2.5}$ concentration during the study period was well above the World Health Organization's $\text{PM}_{2.5}$ ambient air guidelines. Extreme gradient boosting (xgbTree) performed best with an $R^2$ of 0.84.

An air quality monitoring system \cite{r115}, as part of a pilot initiative of Qatar Mobility Innovations Center (QMIC), coupled with support vector machines (SVM), M5P model trees, and artificial neural networks (ANN) were used to predict the concentrations of ground–level ozone $\text{O}_{3}$, nitrogen dioxide $\text{NO}_{2}$, and sulfur dioxide $\text{SO}_{2}$. Results showed that the use of different features in multivariate modeling with M5P algorithm yielded the best forecasting performances, with the lowest RMSE, reaching 31.4, when hydrogen sulfide ($\text{H}_{2}{S}$) is used to predict $\text{SO}_{2}$.

\begin{table}
\caption{Summary of included literature using Low-cost enabled monitoring and machine learning for outdoor air pollution prediction and forecasting}
\begin{tabular}{|p{1.3cm}|p{1cm}|p{1.7cm}|p{1.5cm}|p{2cm}|p{2cm}|c|c|}
\hline
Continent &
  \begin{tabular}[c]{@{}l@{}}Source \\ of Data\end{tabular} &
  Pollutant (s) &
  Method &
  Model (s) &
  \begin{tabular}[c]{@{}l@{}}Deployment\\  Strategy\end{tabular} &
  \multicolumn{1}{l|}{\begin{tabular}[c]{@{}l@{}}Number \\ of Sensors\end{tabular}} &
  \multicolumn{1}{l|}{Reference} \\ \hline
 &
  China &
  $\text{PM}_{2.5}$ &
  Estimation &
  GPR, KD-Tree &
  \begin{tabular}[c]{@{}l@{}}Taxis with \\ built-in sensors \end{tabular} &
  151 &
 \cite{r30} \\ \cline{2-8} 
 &
  China &
  $\text{PM}_{2.5}$, TOVC &
  Estimation &
  RF &
  \begin{tabular}[c]{@{}l@{}}Crowd \\sourcing \\vehicles with \\ built-in sensors\end{tabular} &
  500 &
  \cite{r42} \\ \cline{2-8} 
 &
  China &
  $\text{PM}_{2.5}$ &
  Estimation &
  \begin{tabular}[c]{@{}l@{}}3D-PCEM, LI, \\ MLR, ANN\\ \& GP\end{tabular} &
\begin{tabular}[c]{@{}l@{}}Taxis with \\built-in \\ sensors \& a \\Customized\\  Hexacopter\end{tabular} &
  6 &
  \cite{r45} \\ \cline{2-8} 
 &
  China &
  $\text{PM}_{2.5}$ &
  Forecasting & LSTM &
  Fixed Sensor Nodes &
  1 &
  \cite{r68} \\ \cline{2-8} 
 &
  China &
  $\text{PM}_{2.5}$ &
  Estimation &
  \begin{tabular}[c]{@{}l@{}}Xgboost, RF, \\ K-NN, SVR, \\ ANN.\end{tabular} &
\begin{tabular}[c]{@{}l@{}}Vehicles with\\  built-in sensors\end{tabular} &
  260 &
  \cite{r114} \\ \cline{2-8} 
 &
  Qatar &
  $\text{O}_{3}$, $\text{SO}_{2}$, $\text{NO}_{2}$ &
  Forecasting &
  \begin{tabular}[c]{@{}l@{}}SVM, ANN, \\ M5P Model\\ Trees\end{tabular} &
  Fixed Sensor Nodes &
  1 &
  \cite{r115} \\ \cline{2-8} 
\multirow{-7}{*}{Asia} &
  Korea &
  $\text{PM}_{2.5}$ &
  Forecasting &
  ARIMA, VARMA &
  Fixed Sensor Nodes &
  15 &
  \cite{r55} \\ \hline
 &
  Morocco &
  $\text{PM}_{2.5}$, $\text{PM}_{10}$ &
  Prediction &
  MLR, SVR, RF &
  \begin{tabular}[c]{@{}l@{}}Nomadic \& \\Mobile \\ Sensor Nodes\end{tabular} &
  7 &
  \cite{r83} \\ \cline{2-8} 
 &
  Nigera &
  $\text{PM}_{2.5}$ &
  Forecasting &
  \begin{tabular}[c]{@{}l@{}}ARIMA,\\ Prophet,\\  ES, NNAR, \\SVM, XGBoost, \\RF\end{tabular} &
  Fixed Sensor Nodes &
  1 &
  \cite{r103} \\ \cline{2-8} 
\multirow{-3}{*}{Africa} &
  Uganda &
  $\text{PM}_{2.5}$ &
  Prediction &
  \begin{tabular}[c]{@{}l@{}}XgbTree, RF, \\GAM , KNN , \\RR, OLS, \\ LASSO, PCA\end{tabular} &
  \begin{tabular}[c]{@{}l@{}}Fixed \& Mobile\\  Sensor nodes\end{tabular} &
  80 &
  \cite{r130} \\ \hline
\begin{tabular}[c]{@{}l@{}}South\\  America\end{tabular} &
  Ecuador &
  $\text{PM}_{2.5}$ &
  Prediciton &
  Decision Trees &
  \begin{tabular}[c]{@{}l@{}}Fixed Sensor\\ Nodes \& \\Google Traffic\end{tabular} &
  1 &
  \cite{r75} \\ \hline
 &
  USA &
  $\text{NO}_{2}$, $\text{O}_{3}$ &
  Prediction &
  RF , RF-K &
  Mobile Sensor Nodes &
  31 &
  \cite{r7} \\ \cline{2-8} 
\multirow{-2}{*}{\begin{tabular}[c]{@{}l@{}}North\\  America\end{tabular}} &
  Canada &
  $\text{PM}_{2.5}$, BC &
  Prediction &
  \begin{tabular}[c]{@{}l@{}}ANN, \\  GBM, LUR\end{tabular} &
  Mobile Sensor Nodes &
  2 &
  \cite{r112} \\ \hline
\end{tabular}
\label{fig:LowCostTable}
\end{table}

\subsection{Applications of high-cost monitoring and ML for outdoor air pollution}

\begin{figure}[h]
    \centering
    \includegraphics[width=0.9\textwidth]{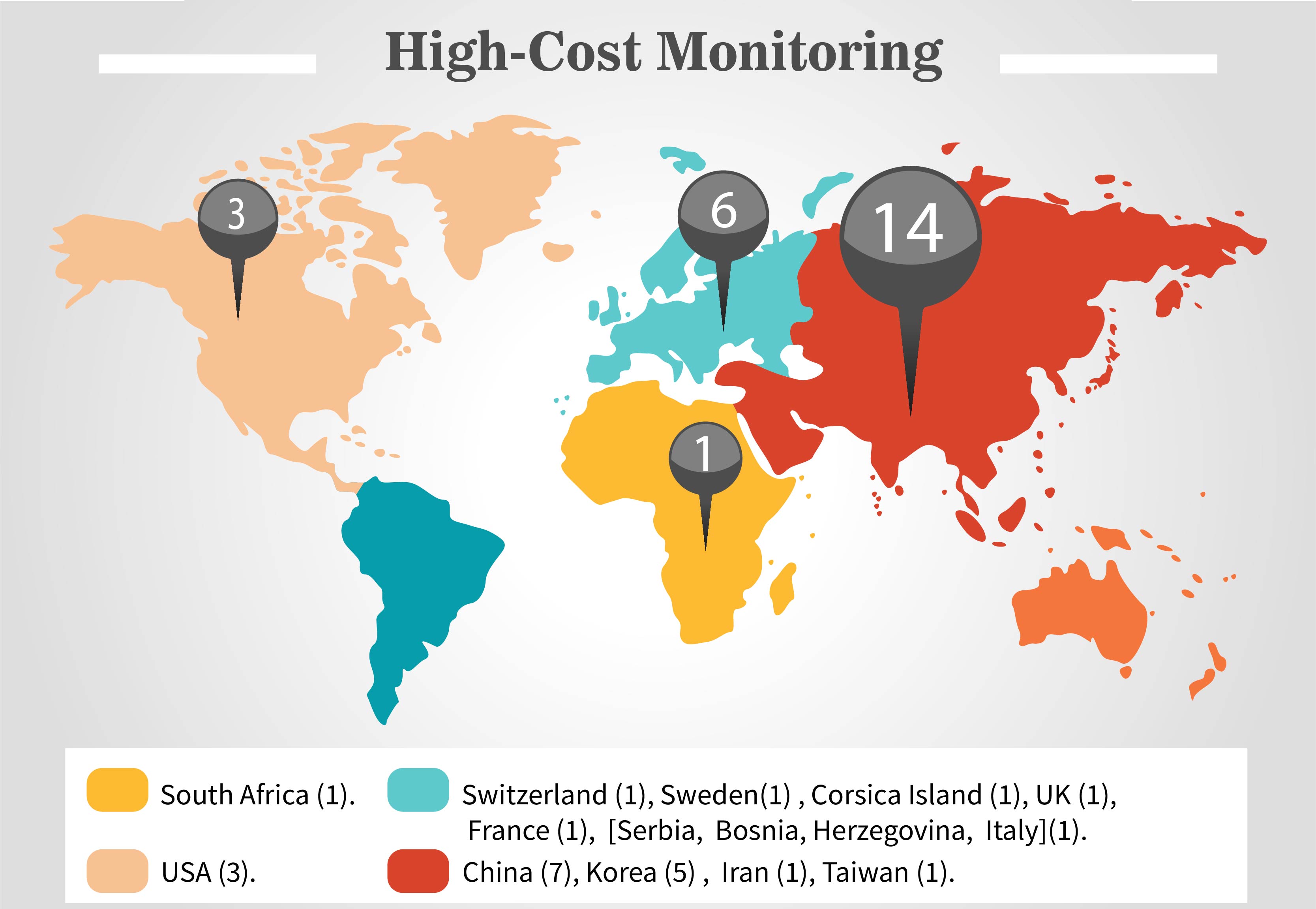}
    \caption{Geographical distribution of included papers using high-cost monitoring stations}
    \label{fig:HCMap}
\end{figure}

\begin{figure}[H]
\centering
\includegraphics[width=\textwidth]{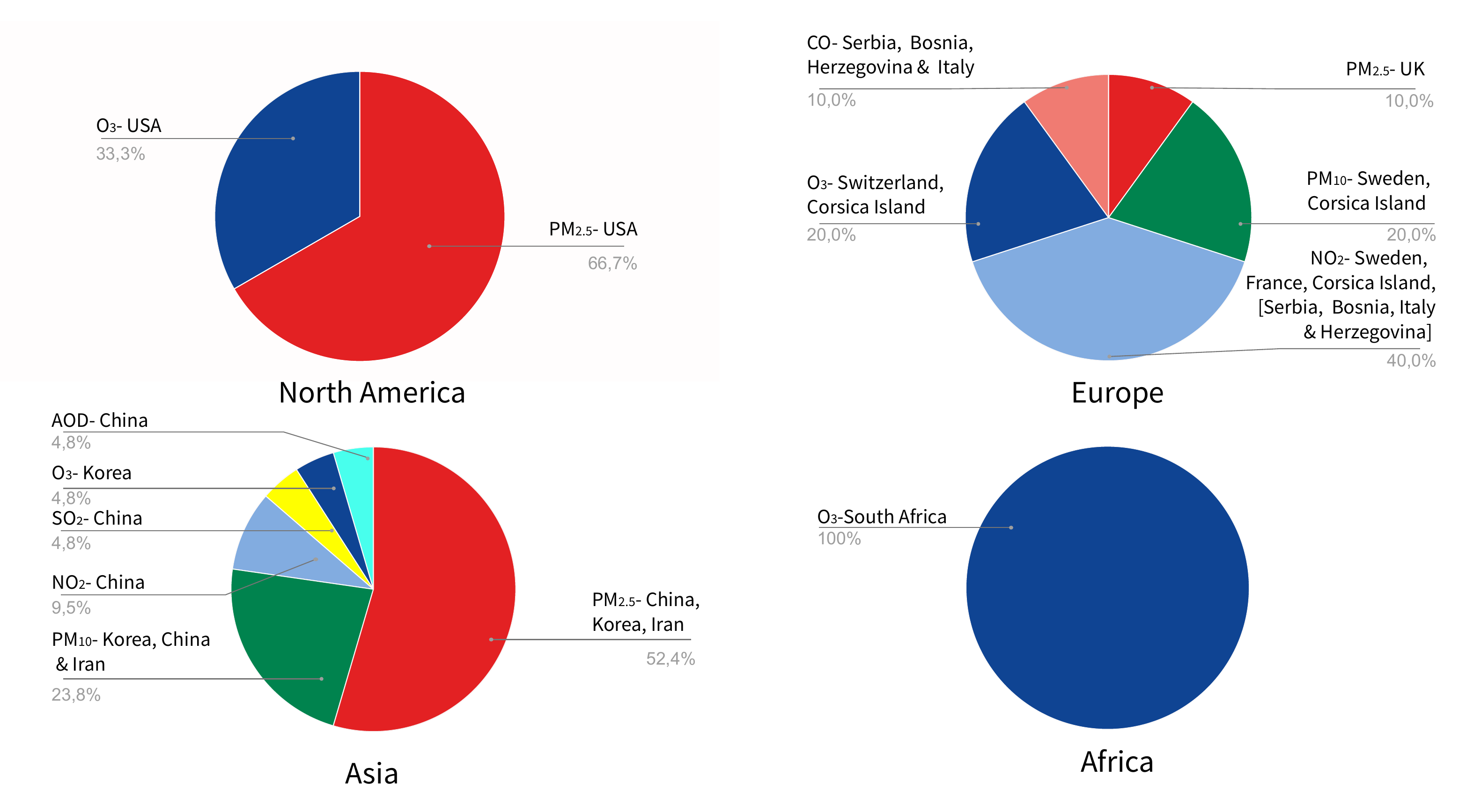}\\
\caption{Proportion of pollutants predicted in included literature using High-cost enabled monitoring and machine learning for outdoor air pollution prediction. \textit{ $\text{PM}_{2.5}$- Particulate matter with a diameter <2.5 microns, $\text{PM}_{10}$- Particulate matter with a diameter <10 microns, $\text{O}_{3}$- Ozone, $\text{SO}_{2}$- Sulfur Dioxide, $\text{NO}_{2}$- Nitrogen Dioxide, $\text{CO}$- Carbon Monoxide,
$\text{AOD}$- Aerosol Optical Depth.}}
\label{fig:AllHC}
\end{figure}

Geographically, selected papers presenting applications using high-cost monitoring stations and machine learning for outdoor air pollution prediction use data collected mostly from Asia, followed by Europe, while only three papers were identified in North America, one in Africa and none in South America (Fig. \ref{fig:HCMap}). 
A wider range of pollutants was predicted based on data from Asia and Europe. As can be seen in Fig. \ref{fig:AllHC},  $\text{PM}_{2.5}$ and $\text{O}_{3}$ were the only pollutants predicted in North America, while in Africa, only $\text{O}_{3}$ measurements were considered. A summary of pollutants, geographical data sources, estimation methods, machine learning models and number of monitoring stations used in the included literature is illustrated in Table \ref{fig:HighCostTable}. A $"*"$ was placed in front of the one paper that was mentioned twice.

\subsubsection{Time series prediction}

Studies conducting time series prediction used pollution data from various regions and countries such as South Korea \cite{r89,r79,r96}, Serbia, Bosnia and Herzegovina, and Italy \cite{r46}, Hong Kong and China \cite{r52,r100} and the United Kingdom \cite{r134}.

$\text{PM}_{2.5}$ concentrations were forecast in \cite{r89} using ML models (XGB, LGBM, GRU, CNNLSTM and BiULSTM) and measurements of pollutants, meteorological, and predictive data from CMAQ models as inputs. Results indicated that LSTM outperformed gradient tree boosting models, recurrent, and convolutional neural networks by scoring the lowest MAE and RMSE for almost all stations. $\text{PM}_{2.5}$ and $\text{PM}_{10}$ were forecast using LSTM and deep auto-encoder methods (DAE) in \cite{r96}, and novel hybrid models combining CNN-LSTM and CNN-GRU in \cite{r79}. These works forecast concentrations of these two pollutants for up to 15 days (\cite{r79}) and 10 days (\cite{r96}). 

Different works forecast various combinations of pollutants. Concentrations of $\text{PM}_{2.5}$, $\text{NO}_{2}$ and $\text{SO}_{2}$ were predicted in \cite{r52}, $\text{PM}_{2.5}$ and $\text{NO}_{x}$ in \cite{r100}, $\text{PM}_{2.5}$, $\text{PM}_{10}$ and $\text{NO}_{x}$ in \cite{r134}, $\text{CO}$ and $\text{NO}_{2}$ in \cite{r46} and $\text{PM}_{10}$ and $\text{NO}_{2}$ in \cite{r27}.  

$\text{PM}_{10}$ and $\text{NO}_{2}$ measurements were predicted using historical hourly air pollutant and meteorology records of over 10 years in \cite{r27}. Results showed improved performance over baseline methods for $\text{NO}_{2}$ predictions, while similar performances were achieved by the deep LSTM model and the proposed framework for $\text{PM}_{10}$ predictions. 
Meteorological features were used for a time series prediction of $\text{NO}_{2}$ and $\text{CO}$ in \cite{r46}. Air pressure, relative humidity, daily temperature, and wind speed were used to train a Nonlinear Autoregressive Exogenous (NARX) neural network, with the best predictors of $\text{NO}_{2}$ found to be air pressure and relative humidity, followed by wind speed, while pressure and temperature (Bosnia and Italy) and wind speed (Serbia) were found to be the best predictors for CO.

Random Forest was used in \cite{r52} to predict the concentrations of $\text{PM}_{2.5}$, $\text{NO}_{2}$ and $\text{SO}_{2}$, 24 hours in advance, with an improvement in accuracy by 11\% for $\text{PM}_{2.5}$, 4.2\%  for $\text{NO}_{2}$, and by 5.2\% for $\text{SO}_{2}$ in comparison with baseline methods.

Machine learning models including Random Forest (RF), Boosted Regression Trees (BRT), Support Vector Machine (SVM), Extreme Gradient Boosting (XGBoost), Generalized Additive Model (GAM), and Cubist were used in \cite{r100} for hourly street-level prediction of $\text{PM}_{2.5}$ and $\text{NO}_{x}$ at high temporal resolution. Results showed that RF outperformed other MLAs with ten-fold cross validation. Additionally, results claimed that non-emission factors contributed 84\% and 65\% to the predictions of street-level $\text{PM}_{2.5}$ and $\text{NO}_{x}$ concentrations, respectively. Non-local pollution and temperature were the major non-emission factors, whereas private cars were the major emission contributor \cite{r100}. 

Similarly to the previously mentioned papers, \cite{r134} reinforced the finding that including more variables as predictors improves upon prediction performance. In \cite{r134}, a regression model, a random forest model and a combination of the two using GAM were used to predict $\text{PM}_{2.5}$ measurements, with inputs being $\text{PM}_{10}$, $\text{NO}_{x}$ measurements and meteorological features. Results of this study showed that the combination of the regression and RF models performed a cross-validation $R^2$ of 99.29\% and a Mean Square Error near 1. 

\subsubsection{Feature-based prediction}
Literature conducting feature based prediction used pollution data from various regions such the United States \cite{r5}, South Korea \cite{r15}, Hangzhou city \cite{r28} and other cities in China \cite{r36}, and France \cite{r146}.
 
Daily $\text{PM}_{2.5}$ measurements from the EPA's Air Quality System (AQS) were used to evaluate and compare the predictive performance of exposure modeling approaches in \cite{r5}.
Factors including season, 'urbanicity', and levels of $\text{PM}_{2.5}$ concentration were included as predictors to test the performance of the different methods, with results showing that the downscaler model and universal kriging outperformed machine learning algorithms regardless of season, urbanicity and $\text{PM}_{2.5}$ concentration levels. 

$\text{PM}_{10}$ concentrations were predicted in \cite{r15} using Extreme Gradient Boosting (XGBoost) and Light Gradient Boosting Machine (LightGBM). Both algorithms used a combination of meteorological, emission rate data and output features of Community Multi-Scale Air Quality (CMAQ). Results showed that XGBoost outperformed LightGBM in terms of error, with an $R^2$ of 0.80, however, in terms of speed, LightGBM was 21.23 times faster.

A semi-supervised learning and pruning method was used in \cite{r28} to infer spatially fine-grained $\text{PM}_{10}$, $\text{PM}_{2.5}$ and $\text{NO}_{2}$ measurements. In addition to air quality data, traffic-related features, road-network-related features, point-of-interest related features, and check-in features extracted from various urban data were also used to realize the estimation. Accuracy of $\text{PM}_{2.5}$, $\text{PM}_{10}$, and $\text{NO}_{2}$ was respectively raised to 0.624, 0.628, and 0.636 from 0.443, 0.446, and 0.494. 

Hourly measurements of meteorological data including temperature, wind speed and direction, humidity and pressure were used to train machine learning methods (Linear Regression, Support Vector Machine, Neural Network, and Random Forest) to predict $\text{PM}_{2.5}$ levels at an hourly timescale in  \cite{r36}. RF achieved the highest predictive accuracy, with a $R^2$ between 0.67 and 0.78. The ANN model was ranked second with a $R^2$ between 0.60 and 0.71, followed by SVR, while Linear Regression achieved the lowest performance. 

Spatial correlation between pollution monitoring stations was leveraged to predict $\text{NO}_{2}$ measurements in \cite{r146}. Results showed that correlation-based station selection outperformed distance-based station selection and provided an $R^2$ above 0.8.

\subsubsection{Spatio-temporal prediction}

$\text{PM}_{2.5}$ concentrations were forecasted for the next 24 hours in \cite{r13} using the geo-context-based diffusion convolutional recurrent neural network, GC-DCRN. The model was evaluated on two real-world air quality datasets (Los Angeles and Beijing AQ datasets), and outperformed baseline models at all forecasting horizons on the Beijing dataset. 

$\text{PM}_{2.5}$ concentrations for up to 48 hours were forecasted in \cite{r56} using multiple neural networks including ANN, CNN, and LSTM to extract spatial-temporal relations. A predictive model including meteorological data and terrain aspect features was used and evaluated on two real world datasets from Taiwan and Beijing, and outperformed state-of-the-art models. 

Spatio-temporal distributions of daily $\text{PM}_{2.5}$ concentrations in China were predicted in \cite{r92} using a geographically-weighted gradient boosting machine (GW-GBM) and resulted in improved results over the original GBM model with an $R^2$ of 0.76. 

Pollutant peaks were detected in Corsica Island using a combination of ANNs and clustering in \cite{r31}. MultiLayer Perceptron was used to forecast hourly concentrations of ozone ($\text{O}_{3}$), nitrogen dioxide ($\text{NO}_{2}$) and particulate matter ($\text{PM}_{10}$) 24 hours ahead, by using other pollutants, meteorological measures and output pollutants as inputs. The model was then hybridized with hierarchical clustering and a combination of self-organizing map and k-means clustering. The results showed that these hybrid models outperformed classical MLP for high concentration prediction of $\text{PM}_{10}$ and $\text{O}_{3}$.

$\text{PM}_{10}$ and $\text{PM}_{2.5}$ pollution concentrations were predicted in \cite{r78} using Support Vector Machine, Geographically Weighted Regression, Artificial Neural Network and Auto-regressive Nonlinear Neural Network. 
Inputs included day of week, month of year, topography, meteorology, and pollutant rate of two nearest neighbors. A prediction model was used to improve these methods by filtering existing noise and predicting missing data in meteorological and air pollution data. As a result, error percentages were improved by 57\%, 47\%, 47\% and 94\%, respectively. Results claimed that the best and most accurate method for predicting air pollution in the city of Tehran is the NARX method with refined data.
\begin{table}
\caption{Summary of included literature using High-cost enabled monitoring and machine learning for outdoor air pollution prediction and forecasting.}
\small{
\begin{tabular}{|p{1.3cm}|p{2cm}|p{1.7cm}|p{1.5cm}|p{3cm}|p{2cm}|l|}
\hline
\multicolumn{1}{|l|}{Continent} &
  Source of Data &
  Pollutant (s) &
  Method &
  Model (s) &
  \multicolumn{1}{l|}{\begin{tabular}[c]{@{}l@{}}Number \\ of monitoring\\stations\end{tabular}} &
  Reference \\ \hline
 &
  China &
  $\text{PM}_{2.5}$, $\text{NO}_{2}$, $\text{SO}_{2}$ &
  Prediction &
  RF &
  1 &
  \cite{r52} \\ \cline{2-7} 
 &
  Taiwan \& China &
  $\text{PM}_{2.5}$ &
  Forcasting &
  ANN, CNN, and LSTM &
  \multicolumn{1}{l|}{-} &
  \cite{r56} \\ \cline{2-7} 
 &
  \begin{tabular}[c]{@{}l@{}}* China \& USA \end{tabular} &
  $\text{PM}_{2.5}$ &
  Forecasting &
  GC-DCRN &
13 &
  \cite{r13} \\ \cline{2-7} 
 &
  Korea &
  $\text{PM}_{10}$ &
  Prediction &
XGBoost, LighGBM &
  299 &
  \cite{r15} \\ \cline{2-7} 
 &
  China &
 $\text{PM}_{2.5}$, $\text{PM}_{10}$, $\text{NO}_{2}$ &
  Estimation & Decision Tree, KNN &
  6 &
  \cite{r28} \\ \cline{2-7} 
 &
  China &
  $\text{PM}_{2.5}$ &
  Prediction &
  LR, SVM, ANN \& RF &
  11 &
  \cite{r36} \\ \cline{2-7} 
 &
  Iran &
  $\text{PM}_{10}$, $\text{PM}_{2.5}$ &
  Prediction &
  \begin{tabular}[c]{@{}l@{}}SVR, GWR,\\ ANN, and NARX\end{tabular} &
  24 &
  \cite{r78} \\ \cline{2-7} 
 &
  Korea &
  $\text{PM}_{10}$, $\text{PM}_{2.5}$ &
  Forecasting &
   CNN, LSTM, GRU &
  39 &
  \cite{r79} \\ \cline{2-7} 
 &
  Korea &
  $\text{PM}_{2.5}$ &
  Estimation &
  \begin{tabular}[c]{@{}l@{}}XGB , LGBM , GRU, \\ CNNLSTM,\\ BiULSTM, LSTM\end{tabular} &
  8 &
  \cite{r89} \\ \cline{2-7} 
 &
  China &
  $\text{PM}_{2.5}$, AOD &
  Prediction &
  GW-GBM &
  \multicolumn{1}{l|}{-} &
  \cite{r92} \\ \cline{2-7} 
 &
  Korea &
  $\text{PM}_{2.5}$,$\text{PM}_{10}$ &
  Forecasting &
  LSTM \& DAE &
  25 &
  \cite{r96} \\ \cline{2-7} 
 &
  Korea &
  $\text{O}_{3}$ &
  Forecasting &
  CNN &
  255 &
  \cite{r10} \\ \cline{2-7} 
\multirow{-13}{*}{Asia} &
  Hong Kong &
  $\text{PM}_{2.5}$&
  Prediction &
  \begin{tabular}[c]{@{}l@{}}RF, BRT, SVM,\\ XGBoost, GAM \\ \& Cubist\end{tabular} &
  3 &
  \cite{r100} \\ \hline
 
 &
  Switzerland &
  $\text{O}_{3}$ &
  Forecasting &
  \begin{tabular}[c]{@{}l@{}}Penalized linear\\ regression algorithms, \\ ARIMAX, RF, \\LSBOOST, XGBOOST, \\NGBOOST, LM, Lasso, \\Ridge\end{tabular} &
  3 &
  \cite{r8} \\ \cline{2-7} 
 &
  Sweden &
  $\text{PM}_{10}$, $\text{NO}_{2}$ &
  Prediction &
  Deep Neural Networks &
  5 &
  \cite{r27} \\ \cline{2-7} 
 &
  Corsica Island &
  $\text{O}_{3}$, $\text{NO}_{2}$, $\text{PM}_{10}$ &
  Forecasting &
  MLP, Self-Organizing map and k-means clustering &
  9 &
  \cite{r31} \\ \cline{2-7} 
 &
  \begin{tabular}[c]{@{}l@{}}Serbia,  Bosnia, \\ Herzegovina, \\ Italy\end{tabular} &
  $\text{NO}_{2}$, $\text{CO}$ &
  Prediction &
  NARX &
  3 &
  \cite{r46} \\ \cline{2-7} 
 &
  UK &
  $\text{PM}_{2.5}$ &
  Prediction &
  RF &
  \multicolumn{1}{l|}{-} &
  \cite{r134} \\ \cline{2-7} 

\multirow{-8}{*}{Europe} &
  France &
   $\text{NO}_{2}$ &
  Estimation &
  MLR &
  20 &
  \cite{r146} \\ \hline
 &
  USA &
  $\text{PM}_{2.5}$ &
  Prediction &
  \begin{tabular}[c]{@{}l@{}}OLS and Inverse\\ distance weighting, \\Kriging, Statistical\\ downscaling models,\\ LUR, NN, RF, \& SVR\end{tabular} &
  829 &
  \cite{r5} \\ \cline{2-7} 
 &
  USA &
  O3 &
  Prediction &
  \begin{tabular}[c]{@{}l@{}}LM, RIDGE, LASSO, \\ELASTICNET, PCR,\\ PLSR, KNN,\\ SVR, BPNN, DNN,\\RT, RF, XBOOST\end{tabular} &
  1313 &
  \cite{r125} \\ \cline{2-7} 
  \multirow{-2}{*}{\begin{tabular}[c]{@{}l@{}}North\\  America\end{tabular}} &

  *China \& USA &
  $\text{PM}_{2.5}$ &
  Forecasting &
  GC-DCRN &
  13 &
  \cite{r13} \\ \hline
Africa &
  South Africa &
  $\text{O}_{3}$ &
  Estimation &
  ANN &
  1 &
  \cite{r97} \\ \hline
\end{tabular}}
\label{fig:HighCostTable}
\end{table}

Ground-level day-ahead maximum ozone concentration was forecast using machine learning in \cite{r8}, while a convolutional neural network (CNN) model was used to forecast hourly ozone concentrations 14 days in advance \cite{r10}. Linear Regression (LR) models and Artificial Neural Networks (ANNs) were applied to predict levels of ground level Ozone at particular locations based on the cross-correlation and spatial-correlation of different air pollutants in South Africa \cite{r97}. Linear Regression had an $R^2$ of 0.579 and identified the most important predictors of Ozone as temperature, relative humidity, nitrogen dioxide and other meteorological features and pollutants. Neural network yielded better results, with an accuracy that reached 79\%. 

Thirteen spatial and spatio-temporal models were compared in \cite{r125} with the aim of estimating daily maximum 8-hour average ambient ozone concentrations across the Contiguous U.S. 
Predictors used included chemistry-transport model predictions, meteorological factors, land use and  cover, stationary and mobile emissions. Results showed that by tuning the sample weights, spatio-temporal models were able to predict concentrations used to calculate ozone design values that perform comparably or better than spatial models (nearly 30\% decrease of cross-validated RMSE).

\subsection{Applications of hybrid monitoring and ML for outdoor air pollution}

Only two papers \cite{r2,r12} from the selected literature used a combination of low-cost IoT and expensive monitoring stations for machine learning techniques to forecast air pollution. 

A combination of 21 fixed stations and 15 mobile air quality sensors deployed in vehicles were used in China  \cite{r2}, with a machine learning based framework (Deep-MAPS) aiming for the spatial inference of $\text{PM}_{2.5}$ concentrations. The ML framework was based on Gradient Boosting Decision Tree (GBDT) and used air quality data along heterogeneous features such as geographical characteristics, land use, transport, public vitality, and meteorological features. Results showed that Deep-MAPS outperformed all other benchmark methods, and that $R^2$ increased and the margin of error decreased with every increment of the percentage of mobile data included. Using fixed-location data only, the model yielded an $R^2$ of 0.70, and reached an $R^2$ of 0.85 by using fixed data and 100\% mobile data.

The second work using hybrid-cost IoT was proposed in \cite{r12}. This study also relied on multi-sourced and heterogeneous data to generate high-resolution air quality maps for cities. The authors of this paper leveraged four different data sources including air quality measurements produced by a mobile sensor network of low-cost nodes installed on top of ten public buses, two high-cost static monitoring stations, land-use and traffic data were leveraged in \cite{r12}. To estimate Lung deposited surface area (LDSA) values, two modeling approaches were used, a Log-linear Regression model and a Deep Learning approach using auto-encoder structures. The second proposed method based on Deep Neural Networks outperformed the other models.

\begin{table}[H]
\caption{Summary of included literature using Hybrid enabled monitoring and machine learning for outdoor air pollution prediction and forecasting}
{\small
\begin{tabular}{llllllll}
\hline
\multicolumn{1}{|l|}{Continent} &
  \multicolumn{1}{l|}{\begin{tabular}[c]{@{}l@{}}Source\\  of Data\end{tabular}} &
  \multicolumn{1}{l|}{Pollutant (s)} &
  \multicolumn{1}{l|}{Method} &
  \multicolumn{1}{l|}{Model (s)} &
  \multicolumn{1}{l|}{\begin{tabular}[c]{@{}l@{}}Deployment \\ Strategy\end{tabular}} &
  \multicolumn{1}{l|}{\begin{tabular}[c]{@{}l@{}}Number\\  of Sensors\end{tabular}} &
  \multicolumn{1}{l|}{Reference} \\ \hline
\multicolumn{1}{|l|}{Asia} &
  \multicolumn{1}{l|}{\cellcolor[HTML]{FFFFFF}{\color[HTML]{222222} China}} &
  \multicolumn{1}{l|}{\cellcolor[HTML]{FFFFFF}{\color[HTML]{222222} $\text{PM}_{2.5}$}} &
  \multicolumn{1}{l|}{\cellcolor[HTML]{FFFFFF}{\color[HTML]{222222} Inference}} &
  \multicolumn{1}{l|}{\cellcolor[HTML]{FFFFFF}\begin{tabular}[c]{@{}l@{}}Gradient\\ Boosting \\ Decision Tree\end{tabular}} &
  \multicolumn{1}{l|}{\cellcolor[HTML]{FFFFFF}\begin{tabular}[c]{@{}l@{}}Fixed Stations \& \\ Mobile Sensor\\ nodes\end{tabular}} &
  \multicolumn{1}{c|}{\cellcolor[HTML]{FFFFFF}36} &
  \multicolumn{1}{c|}{\cellcolor[HTML]{FFFFFF}\cite{r2}} \\ \hline
\multicolumn{1}{|l|}{Europe} &
  \multicolumn{1}{l|}{\cellcolor[HTML]{FFFFFF}{\color[HTML]{222222} Switzerland}} &
  \multicolumn{1}{l|}{\cellcolor[HTML]{FFFFFF}LDSA} &
  \multicolumn{1}{l|}{\cellcolor[HTML]{FFFFFF}Estimation} &
  \multicolumn{1}{l|}{\cellcolor[HTML]{FFFFFF}{\color[HTML]{222222} \begin{tabular}[c]{@{}l@{}}LRM, BLL, \\NLL, BLL-LU\end{tabular}}} &
  \multicolumn{1}{l|}{\cellcolor[HTML]{FFFFFF}\begin{tabular}[c]{@{}l@{}}Fixed Stations \&\\  Sensing nodes\\ anchored to\\ public buses.\end{tabular}} &
  \multicolumn{1}{c|}{\cellcolor[HTML]{FFFFFF}12} &
  \multicolumn{1}{c|}{\cellcolor[HTML]{FFFFFF}\cite{r12}} \\ \hline
 &  &  &  &  &  &  &  \\
 &  &  &  &  &  &  &  \\
 &  &  &  &  &  &  &  \\
 &  &  &  &  &  &  & 
\end{tabular}}
\label{HybridTable}
\end{table}

A summary of pollutants, geographical data sources, estimation methods, machine learning models, number of sensors and deployment strategies used in the included literature is illustrated in Table \ref{HybridTable}.

Table \ref{bigtable} presents a summary of the combinations of low-cost, high-cost and hybrid monitoring methods and input features used for outdoor air pollution prediction and forecasting in selected papers of the systematic literature review. The column "other pollutants" indicates whether the work used some pollutants to infer the values of others. For instance, in \cite{r12}, the authors used $\text{CO}$, $\text{NO}_{2}$, $\text{SO}_{2}$, and $\text{VOC}$ (benzene, toluene, ethyl benzene, M + P xylene, O-xylene)) as features to predict the target variable $\text{PM}_{2.5}$. On the other hand, in \cite{r30} no other pollutants were used to predicts PM2.5, they rather used meteorological features and location as predictors. 

\begin{table}
\caption{Summary of various combinations of monitoring sensors and input features used for outdoor air pollution prediction and forecasting in included literature}
\small \begin{tabular}{|
>{\columncolor[HTML]{FFFFFF}}c |
>{\columncolor[HTML]{FFFFFF}}l |
>{\columncolor[HTML]{FFFFFF}}l |
>{\columncolor[HTML]{FFFFFF}}l |
>{\columncolor[HTML]{FFFFFF}}l |
>{\columncolor[HTML]{FFFFFF}}l |
>{\columncolor[HTML]{FFFFFF}}l |
>{\columncolor[HTML]{FFFFFF}}l |
>{\columncolor[HTML]{FFFFFF}}l |}
\hline
\multicolumn{1}{|l|}{\cellcolor[HTML]{FFFFFF}{\color[HTML]{000000} Category}} &
  {\color[HTML]{000000} Reference} &
  \multicolumn{1}{c|}{\cellcolor[HTML]{FFFFFF}{\color[HTML]{000000} \begin{tabular}[c]{@{}c@{}}Other \\ pollutants\end{tabular}}} &
  \multicolumn{1}{c|}{\cellcolor[HTML]{FFFFFF}{\color[HTML]{000000} \begin{tabular}[c]{@{}c@{}}Meteorological \\ features\end{tabular}}} &
  \multicolumn{1}{c|}{\cellcolor[HTML]{FFFFFF}{\color[HTML]{000000} \begin{tabular}[c]{@{}c@{}}Traffic\\ related \\ features\end{tabular}}} &
  \multicolumn{1}{c|}{\cellcolor[HTML]{FFFFFF}{\color[HTML]{000000} \begin{tabular}[c]{@{}c@{}}Location\\  (Longitude \\ \& Latitude)\end{tabular}}} &
  \multicolumn{1}{c|}{\cellcolor[HTML]{FFFFFF}{\color[HTML]{000000} \begin{tabular}[c]{@{}c@{}}Land-use \\ Features\end{tabular}}} &
  \multicolumn{1}{c|}{\cellcolor[HTML]{FFFFFF}{\color[HTML]{000000} \begin{tabular}[c]{@{}c@{}}Context\\ Specific\\  features\end{tabular}}} &
  \multicolumn{1}{c|}{\cellcolor[HTML]{FFFFFF}{\color[HTML]{000000} \begin{tabular}[c]{@{}c@{}}Population\\  density\end{tabular}}} \\ \hline
\cellcolor[HTML]{FFFFFF}{\color[HTML]{000000} } &
  {\color[HTML]{000000} \cite{r30}} &
  {\color[HTML]{000000} } &
  {\color[HTML]{000000} x} &
  {\color[HTML]{000000} } &
  {\color[HTML]{000000} x} &
  {\color[HTML]{000000} } &
  {\color[HTML]{000000} } &
  {\color[HTML]{000000} } \\ \cline{2-9} 
\cellcolor[HTML]{FFFFFF}{\color[HTML]{000000} } &
  {\color[HTML]{000000} \cite{r42}} &
  {\color[HTML]{000000} } &
  {\color[HTML]{000000} x} &
  {\color[HTML]{000000} } &
  {\color[HTML]{000000} } &
  {\color[HTML]{000000} } &
  {\color[HTML]{000000} } &
  {\color[HTML]{000000} } \\ \cline{2-9} 
\cellcolor[HTML]{FFFFFF}{\color[HTML]{000000} } &
  {\color[HTML]{000000} \cite{r45}} &
  {\color[HTML]{000000} } &
  {\color[HTML]{000000} } &
  {\color[HTML]{000000} } &
  {\color[HTML]{000000} x} &
  {\color[HTML]{000000} } &
  {\color[HTML]{000000} } &
  {\color[HTML]{000000} } \\ \cline{2-9} 
\cellcolor[HTML]{FFFFFF}{\color[HTML]{000000} } &
  {\color[HTML]{000000} \cite{r68}} &
  {\color[HTML]{000000} } &
  {\color[HTML]{000000} } &
  {\color[HTML]{000000} } &
  {\color[HTML]{000000} } &
  {\color[HTML]{000000} } &
  {\color[HTML]{000000} } &
  {\color[HTML]{000000} } \\ \cline{2-9} 
\cellcolor[HTML]{FFFFFF}{\color[HTML]{000000} } &
  {\color[HTML]{000000} \cite{r114}} &
  {\color[HTML]{000000} } &
  {\color[HTML]{000000} x} &
  {\color[HTML]{000000} x} &
  {\color[HTML]{000000} x} &
  {\color[HTML]{000000} x} &
  {\color[HTML]{000000} } &
  {\color[HTML]{000000} } \\ \cline{2-9} 
\cellcolor[HTML]{FFFFFF}{\color[HTML]{000000} } &
  {\color[HTML]{000000} \cite{r115}} &
  {\color[HTML]{000000} x} &
  {\color[HTML]{000000} x} &
  {\color[HTML]{000000} } &
  {\color[HTML]{000000} } &
  {\color[HTML]{000000} } &
  {\color[HTML]{000000} } &
  {\color[HTML]{000000} } \\ \cline{2-9} 
\cellcolor[HTML]{FFFFFF}{\color[HTML]{000000} } &
  {\color[HTML]{000000} \cite{r55}} &
  {\color[HTML]{000000} } &
  {\color[HTML]{000000} } &
  {\color[HTML]{000000} } &
  {\color[HTML]{000000} } &
  {\color[HTML]{000000} } &
  {\color[HTML]{000000} } &
  {\color[HTML]{000000} } \\ \cline{2-9} 
\cellcolor[HTML]{FFFFFF}{\color[HTML]{000000} } &
  {\color[HTML]{000000} \cite{r83}} &
  {\color[HTML]{000000} } &
  {\color[HTML]{000000} x} &
  {\color[HTML]{000000} x} &
  {\color[HTML]{000000} } &
  {\color[HTML]{000000} } &
  {\color[HTML]{000000} } &
  {\color[HTML]{000000} } \\ \cline{2-9} 
\cellcolor[HTML]{FFFFFF}{\color[HTML]{000000} } &
  {\color[HTML]{000000} \cite{r103}} &
  {\color[HTML]{000000} } &
  {\color[HTML]{000000} x} &
  {\color[HTML]{000000} } &
  {\color[HTML]{000000} } &
  {\color[HTML]{000000} } &
  {\color[HTML]{000000} } &
  {\color[HTML]{000000} } \\ \cline{2-9} 
\cellcolor[HTML]{FFFFFF}{\color[HTML]{000000} } &
  {\color[HTML]{000000} \cite{r130}} &
  {\color[HTML]{000000} } &
  {\color[HTML]{000000} x} &
  {\color[HTML]{000000} x} &
  {\color[HTML]{000000} x} &
  {\color[HTML]{000000} x} &
  {\color[HTML]{000000} x} &
  {\color[HTML]{000000} x} \\ \cline{2-9} 
\cellcolor[HTML]{FFFFFF}{\color[HTML]{000000} } &
  {\color[HTML]{000000} \cite{r75}} &
  {\color[HTML]{000000} } &
  {\color[HTML]{000000} x} &
  {\color[HTML]{000000} x} &
  {\color[HTML]{000000} } &
  {\color[HTML]{000000} } &
  {\color[HTML]{000000} } &
  {\color[HTML]{000000} } \\ \cline{2-9} 
\cellcolor[HTML]{FFFFFF}{\color[HTML]{000000} } &
  {\color[HTML]{000000} \cite{r7}} &
  {\color[HTML]{000000} } &
  {\color[HTML]{000000} } &
  {\color[HTML]{000000} x} &
  {\color[HTML]{000000} x} &
  {\color[HTML]{000000} x} &
  {\color[HTML]{000000} } &
  {\color[HTML]{000000} } \\ \cline{2-9} 
\multirow{-13}{*}{\cellcolor[HTML]{FFFFFF}{\color[HTML]{000000} Low Cost}} &
  {\color[HTML]{000000} \cite{r112}} &
  {\color[HTML]{000000} } &
  {\color[HTML]{000000} x} &
  {\color[HTML]{000000} x} &
  {\color[HTML]{000000} x} &
  {\color[HTML]{000000} x} &
  {\color[HTML]{000000} } &
  {\color[HTML]{000000} } \\ \hline
\cellcolor[HTML]{FFFFFF}{\color[HTML]{000000} } &
  {\color[HTML]{000000} \cite{r5}} &
  {\color[HTML]{000000} } &
  {\color[HTML]{000000} x} &
  {\color[HTML]{000000} } &
  {\color[HTML]{000000} } &
  {\color[HTML]{000000} x} &
  {\color[HTML]{000000} } &
  {\color[HTML]{000000} } \\ \cline{2-9} 
\cellcolor[HTML]{FFFFFF}{\color[HTML]{000000} } &
  {\color[HTML]{000000} \cite{r8}} &
  {\color[HTML]{000000} x} &
  {\color[HTML]{000000} x} &
  {\color[HTML]{000000} } &
  {\color[HTML]{000000} x} &
  {\color[HTML]{000000} } &
  {\color[HTML]{000000} } &
  {\color[HTML]{000000} } \\ \cline{2-9} 
\cellcolor[HTML]{FFFFFF}{\color[HTML]{000000} } &
  {\color[HTML]{000000} \cite{r10}} &
  {\color[HTML]{000000} x} &
  {\color[HTML]{000000} x} &
  {\color[HTML]{000000} } &
  {\color[HTML]{000000} x} &
  {\color[HTML]{000000} } &
  {\color[HTML]{000000} } &
  {\color[HTML]{000000} } \\ \cline{2-9} 
\cellcolor[HTML]{FFFFFF}{\color[HTML]{000000} } &
  {\color[HTML]{000000} \cite{r13}} &
  {\color[HTML]{000000} } &
  {\color[HTML]{000000} x} &
  {\color[HTML]{000000} x} &
  {\color[HTML]{000000} x} &
  {\color[HTML]{000000} x} &
  {\color[HTML]{000000} x} &
  {\color[HTML]{000000} } \\ \cline{2-9} 
\cellcolor[HTML]{FFFFFF}{\color[HTML]{000000} } &
  {\color[HTML]{000000} \cite{r15}} &
  {\color[HTML]{000000} x} &
  {\color[HTML]{000000} x} &
  {\color[HTML]{000000} } &
  {\color[HTML]{000000} } &
  {\color[HTML]{000000} } &
  {\color[HTML]{000000} } &
  {\color[HTML]{000000} } \\ \cline{2-9} 
\cellcolor[HTML]{FFFFFF}{\color[HTML]{000000} } &
  {\color[HTML]{000000} \cite{r27}} &
  {\color[HTML]{000000} } &
  {\color[HTML]{000000} x} &
  {\color[HTML]{000000} } &
  {\color[HTML]{000000} } &
  {\color[HTML]{000000} } &
  {\color[HTML]{000000} } &
  {\color[HTML]{000000} } \\ \cline{2-9} 
\cellcolor[HTML]{FFFFFF}{\color[HTML]{000000} } &
  {\color[HTML]{000000} \cite{r28}} &
  {\color[HTML]{000000} } &
  {\color[HTML]{000000} } &
  {\color[HTML]{000000} x} &
  {\color[HTML]{000000} } &
  {\color[HTML]{000000} } &
  {\color[HTML]{000000} } &
  {\color[HTML]{000000} } \\ \cline{2-9} 
\cellcolor[HTML]{FFFFFF}{\color[HTML]{000000} } &
  {\color[HTML]{000000} \cite{r31}} &
  {\color[HTML]{000000} x} &
  {\color[HTML]{000000} x} &
  {\color[HTML]{000000} } &
  {\color[HTML]{000000} } &
  {\color[HTML]{000000} } &
  {\color[HTML]{000000} } &
  {\color[HTML]{000000} } \\ \cline{2-9} 
\cellcolor[HTML]{FFFFFF}{\color[HTML]{000000} } &
  {\color[HTML]{000000} \cite{r36}} &
  {\color[HTML]{000000} } &
  {\color[HTML]{000000} x} &
  {\color[HTML]{000000} } &
  {\color[HTML]{000000} } &
  {\color[HTML]{000000} } &
  {\color[HTML]{000000} } &
  {\color[HTML]{000000} } \\ \cline{2-9} 
\cellcolor[HTML]{FFFFFF}{\color[HTML]{000000} } &
  {\color[HTML]{000000} \cite{r46}} &
  {\color[HTML]{000000} } &
  {\color[HTML]{000000} x} &
  {\color[HTML]{000000} } &
  {\color[HTML]{000000} } &
  {\color[HTML]{000000} } &
  {\color[HTML]{000000} } &
  {\color[HTML]{000000} } \\ \cline{2-9} 
\cellcolor[HTML]{FFFFFF}{\color[HTML]{000000} } &
  {\color[HTML]{000000} \cite{r52}} &
  {\color[HTML]{000000} x} &
  {\color[HTML]{000000} x} &
  {\color[HTML]{000000} } &
  {\color[HTML]{000000} } &
  {\color[HTML]{000000} } &
  {\color[HTML]{000000} } &
  {\color[HTML]{000000} } \\ \cline{2-9} 
\cellcolor[HTML]{FFFFFF}{\color[HTML]{000000} } &
  {\color[HTML]{000000} \cite{r56}} &
  {\color[HTML]{000000} } &
  {\color[HTML]{000000} x} &
  {\color[HTML]{000000} } &
  {\color[HTML]{000000} x} &
  {\color[HTML]{000000} } &
  {\color[HTML]{000000} } &
  {\color[HTML]{000000} } \\ \cline{2-9} 
\cellcolor[HTML]{FFFFFF}{\color[HTML]{000000} } &
  {\color[HTML]{000000} \cite{r78}} &
  {\color[HTML]{000000} x} &
  {\color[HTML]{000000} x} &
  {\color[HTML]{000000} } &
  {\color[HTML]{000000} x} &
  {\color[HTML]{000000} x} &
  {\color[HTML]{000000} } &
  {\color[HTML]{000000} } \\ \cline{2-9} 
\cellcolor[HTML]{FFFFFF}{\color[HTML]{000000} } &
  {\color[HTML]{000000} \cite{r79}} &
  {\color[HTML]{000000} x} &
  {\color[HTML]{000000} x} &
  {\color[HTML]{000000} } &
  {\color[HTML]{000000} } &
  {\color[HTML]{000000} } &
  {\color[HTML]{000000} } &
  {\color[HTML]{000000} } \\ \cline{2-9} 
\cellcolor[HTML]{FFFFFF}{\color[HTML]{000000} } &
  {\color[HTML]{000000} \cite{r89}} &
  {\color[HTML]{000000} } &
  {\color[HTML]{000000} x} &
  {\color[HTML]{000000} } &
  {\color[HTML]{000000} } &
  {\color[HTML]{000000} } &
  {\color[HTML]{000000} } &
  {\color[HTML]{000000} } \\ \cline{2-9} 
\cellcolor[HTML]{FFFFFF}{\color[HTML]{000000} } &
  {\color[HTML]{000000}\cite{ r92}} &
  {\color[HTML]{000000} x} &
  {\color[HTML]{000000} x} &
  {\color[HTML]{000000} } &
  {\color[HTML]{000000} x} &
  {\color[HTML]{000000} } &
  {\color[HTML]{000000} } &
  {\color[HTML]{000000} } \\ \cline{2-9} 
\cellcolor[HTML]{FFFFFF}{\color[HTML]{000000} } &
  {\color[HTML]{000000} \cite{r96}} &
  {\color[HTML]{000000} } &
  {\color[HTML]{000000} x} &
  {\color[HTML]{000000} } &
  {\color[HTML]{000000} } &
  {\color[HTML]{000000} } &
  {\color[HTML]{000000} } &
  {\color[HTML]{000000} } \\ \cline{2-9} 
\cellcolor[HTML]{FFFFFF}{\color[HTML]{000000} } &
  {\color[HTML]{000000} \cite{r97}} &
  {\color[HTML]{000000} } &
  {\color[HTML]{000000} x} &
  {\color[HTML]{000000} } &
  {\color[HTML]{000000} x} &
  {\color[HTML]{000000} } &
  {\color[HTML]{000000} } &
  {\color[HTML]{000000} } \\ \cline{2-9} 
\cellcolor[HTML]{FFFFFF}{\color[HTML]{000000} } &
  {\color[HTML]{000000} \cite{r100}} &
  {\color[HTML]{000000} } &
  {\color[HTML]{000000} x} &
  {\color[HTML]{000000} x} &
  {\color[HTML]{000000} } &
  {\color[HTML]{000000} } &
  {\color[HTML]{000000} } &
  {\color[HTML]{000000} } \\ \cline{2-9} 
\cellcolor[HTML]{FFFFFF}{\color[HTML]{000000} } &
  {\color[HTML]{000000} \cite{r125}} &
  {\color[HTML]{000000} x} &
  {\color[HTML]{000000} x} &
  {\color[HTML]{000000} } &
  {\color[HTML]{000000} x} &
  {\color[HTML]{000000} x} &
  {\color[HTML]{000000} } &
  {\color[HTML]{000000} } \\ \cline{2-9} 
\multirow{-20}{*}{\cellcolor[HTML]{FFFFFF}{\color[HTML]{000000} High Cost}} &
  {\color[HTML]{000000} \cite{r134}} &
  {\color[HTML]{000000} x} &
  {\color[HTML]{000000} x} &
  {\color[HTML]{000000} } &
  {\color[HTML]{000000} } &
  {\color[HTML]{000000} } &
  {\color[HTML]{000000} } &
  {\color[HTML]{000000} } \\ \hline
\cellcolor[HTML]{FFFFFF}{\color[HTML]{000000} } &
  {\color[HTML]{000000} \cite{r2}} &
  {\color[HTML]{000000} x} &
  {\color[HTML]{000000} x} &
  {\color[HTML]{000000} x} &
  {\color[HTML]{000000} x} &
  {\color[HTML]{000000} x} &
  {\color[HTML]{000000} x} &
  {\color[HTML]{000000} x} \\ \cline{2-9} 
\multirow{-2}{*}{\cellcolor[HTML]{FFFFFF}{\color[HTML]{000000} Hybrid}} &
  {\color[HTML]{000000} \cite{r12}} &
  {\color[HTML]{000000} x} &
  {\color[HTML]{000000} x} &
  {\color[HTML]{000000} x} &
  {\color[HTML]{000000} x} &
  {\color[HTML]{000000} x} &
  {\color[HTML]{000000} x} &
  {\color[HTML]{000000} x} \\ \hline
\end{tabular}
\label{bigtable}
\end{table}

\section{Discussion}\label{discussion}

In this systematic literature review, we formulated two research questions in order to identify the applications of machine learning and IoT for outdoor air pollution forecasting and prediction.
Based on the findings of this review, a cost-based analysis was conducted in order to differentiate between applications using low-cost sensor nodes, high-cost monitoring stations and hybrid methods. This difference was noted to correspond to country income levels. High-cost monitoring stations were found to be more used in developed and rich countries than in developing countries. Synthesis by modeling type was conducted: time series prediction, feature-based prediction and spatio-temporal forecasting. Although time-series prediction offers the advantage of predicting in time, feature-based prediction gives a more detailed view on impacting factors and would be more adequate for spatial interpolation. Spatio-temporal forecasting offers a holistic approach to air pollution forecasting, as it includes both spatial and temporal features. 

It's worth mentioning that this paper focuses mainly on physical sensors. However, there are numerous other approaches that have been recently attempting to address the same issue. For predicting air quality, crowd-sensing, satellite imagery, and social networks are some of the most popular methods. In \cite{LiangLiu}, crowd-sensing and web crawling were both made use of to construct large scale datasets of mobile phone-captured outdoor photos, meteorological data, and air pollution data. On the other hand, instead of phone-captured photos, \cite{wu} employed a mobile crowd-sensing framework, which can identify various air contaminants, to monitor the microscale air quality on shared bikes. Machine learning and deep learning models were deployed afterward to provide end-to-end frameworks for the air quality inference. By utilizing readily available satellite imagery data, many studies \cite{Schneider,El-Nadry,Shelton}, provided machine learning models for the inference and prediction of air pollution concentrations and air quality measures. Different social media platforms, including Twitter \cite{Jyun-Yu,Khaefi}, Sina Weibo \cite{Yan-dong,Zhai}, and Instagram \cite{Jyun-Yu,Khaefi} are used as information sources to monitor air quality of a city. Data from these platforms are either used alone or fused with insights from meteorological, geographical and sometimes satellite data to forecast air quality levels in different areas, or enhance the air quality prediction. The data gathered from social networks is usually handled using natural language processing techniques, while the inference models are based on machine learning or deep learning methods.

Although the findings allowed for a detailed synthesis of applications of machine learning and IoT in outdoor air pollution prediction, a few limitations and gaps were identified, highlighting areas to be further studied and opportunities for practical implications.

\paragraph{\textbf{Identified Gaps}}
Based on the findings of this systematic literature review, several gaps were identified in the applications found in the literature. 
\begin{itemize}
    \item \textit{Lack of coverage}: Although the use of low-cost sensors is a cost-effective method to monitor air pollution, the low numbers used in some works remain a major limitation, and need to be improved. The same limitation can be stated for sophisticated, high cost stations. A balance between maximizing coverage and minimizing cost needs to be established in order to allow for a continuous monitoring of air pollution without loss in accuracy. Hybrid methods thus need to be further studied and applied.
    \item \textit{Lack of diversity of data}: Most works in the literature use data measured in Europe and in Asia, specifically in China. This is highly beneficial in monitoring industrial emissions and curbing the pollution crisis. However, more countries need to be involved in pollution monitoring. Not enough data is available for the continent of Africa. Only a few works were identified from select countries, namely: South Africa, Morocco, Nigeria, and Uganda. Diversity in data is key to conducting further research and improving upon the quality of life and of air worldwide. 
    \item \textit{Lack of diversity of features}: Most works use meteorological data as features for prediction models, leading to a lack of research into other impacting features such as land-use features and context-specific features.
\end{itemize}

\paragraph{\textbf{Directions for future research}}

\begin{itemize}
    \item \textit{Combination of chemical modeling and machine learning}: leveraging machine learning techniques in combination with chemical modeling would allow for the prediction of both sources of pollution and the degree of exposure variation for different areas depending on the distance from initial pollution sources. This could greatly improve preparedness and assist in mitigation efforts. To implement this, we propose the collection of historical pollution data and chemical composition data, and then train machine learning models to match patterns in chemical data with pollution sources, and apply these models to new data for predictions.
    \item \textit{Geographical Comparisons}: Comparative studies on similarities in context-specific features for geographically or demographically similar countries would allow for the development of context-specific models and improve re-usability. To implement this, we propose the selection of a set of geographically or demographically similar countries, where relevant data would be collected, standardized and then analyzed to identify common features and patterns, and used to build context-specific models.
    \item \textit{Spatio-temporal forecasting}: More efforts need to be focused on spatio-temporal forecasting of air pollution, since the spatial component allows for predictions at larger scale. To implement this, we propose the development of a model that incorporates both spatial and temporal factors, integration of real-time sensor data for accurate predictions, and the validation of the model's performance over time.
    \item \textit{Machine learning for cost-effective estimation}: Exposure to pollutants other than PMs can be just as nefarious. Some of these pollutants are complicated to measure and require expensive sensor calibration and maintenance for accurate measurements. Machine learning techniques could be used to infer these measurements and thus allow for improved protection efforts. To implement this, we propose the identification of the pollutants that are expensive or complex to measure directly, collection of data on pollutants that are most correlated to them in the atmosphere where available, development of machine learning algorithms to estimate their concentrations based on available data, and the continuous validation and refining of the models.
    \item \textit{Medical applications}: The combination of air quality monitoring with health data could allow the medical and research communities to find associations between respiratory and cognitive diseases and different pollution sources, and thus assist in providing an improved and personalized care for patients with various diseases. To implement this, we propose the establishment of data sharing agreements between air quality monitoring agencies and healthcare providers, the combination of air quality data with anonymized electronic patient health records, the use of data mining and statistical methods to identify associations, as well as the collaboration with medical professionals to develop personalized care plans.
\end{itemize}

\paragraph{\textbf{Practical Implications}}

\begin{itemize}
    \item \textit{Air Quality Monitoring:} Official entities charged with air quality monitoring need to incorporate machine learning applications into their processes and strive to offer real-time forecasting in diversified areas. Additionally, further efforts need to be invested in uncovering the most influential features in particular regions and environments.
    \item \textit{Global Synergy:} Opportunities for collaboration should be encouraged between developed and developing countries to exchange experiences with the use of high cost and low cost sensors. Efforts to develop and use hybrid methods would improve coverage of air pollution monitoring and prediction globally. 
    \item \textit{Health:} Patients with pollution-related diseases could benefit from continuous monitoring of the peaks and sources of pollution. Doctors and medical researchers would benefit from pollution-related data in association studies and gene-environment interactions research.
    \item \textit{Decision-making:} Various industries linked to air polluting emissions should be more involved in studying distances travelled by pollutants in different scenarios and aim to minimize their impact.
    \item \textit{Urban Planning:} Rapidly expanding cities need to be aware of polluted air zones and recognize areas fit to house residential areas, amenities, services, hospitals, and schools.
    \item \textit{Smart Cities:} Smart homes would benefit from built-in sensors to detect pollution in the surrounding area and inform on air quality levels.
\end{itemize}

\paragraph{\textbf{Limitations}}
The conclusions of this review should be considered in light of a number of limitations. The data sources used in this review did not cover all scientific databases and therefore cannot generalize findings to the entirety of the literature. The scope of the review focused on specific aspects related to physical sensors, and so does not provide a general overview that includes aspects related to social sensors, crowdsourcing, or satellites. Although the process of data extraction and analysis was undertaken with extreme diligence, there is potential for bias.



\section{Conclusion}\label{conclusion}

Over the past decade, a growing body of scientific evidence has linked air pollution to various health issues, both cognitive and respiratory, thus highlighting the importance of air pollution monitoring and forecasting. 
This systematic review incorporates the latest scientific evidence and methods using machine learning and Iot enabled monitoring. Three main types of monitoring are identified: low-cost, high-cost and hybrid. A geographical mapping of monitoring types, impacting features, machine learning models, and pollutants is conducted.
The findings of the review indicate that the combination of hybrid monitoring with machine learning models offers an optimized and cost-effective method to monitor and forecast air pollution. Additionally, the inclusion of diverse factors such as meteorological features, traffic, land-use, context-specific features and population density would positively impact prediction, estimation and forecasting results, as well as improve spatial coverage. This review also sheds light on the major gaps in the literature and proposes directions for future research as well as practical implications in areas such as health, smart cities, and urban planning.




\funding{The work presented in this paper was carried out within the MoreAir project, which is funded by the Belgium Ministry of cooperation through the VLIR UOS programme under grant MA2017TEA446A101.}

\acknowledgments{ We thank Google AI for their support, by providing a Google Africa PhD fellowship to the first author of this paper. }

\conflictsofinterest{ The authors declare no conflict of interest. The funders had no role in the design of the study.}


\reftitle{References}





\end{document}